\pdfoutput=1

\documentclass[11pt]{article}

\usepackage[final]{acl}

\usepackage{times}
\usepackage{latexsym}
\usepackage{arydshln}
\usepackage{multirow}
\usepackage[table]{xcolor}
\usepackage{float}
\usepackage{booktabs}
\usepackage{longtable}
\usepackage{caption}
\usepackage{makecell}
\usepackage{xcolor}
\definecolor{nicegreen}{RGB}{40, 164, 40}
\definecolor{nicered}{RGB}{199, 35, 40}
\definecolor{niceblue}{RGB}{22, 48, 194}
\usepackage[T1]{fontenc}

\usepackage[utf8]{inputenc}

\usepackage{stfloats}
\usepackage{hyperref}

\usepackage{microtype}

\usepackage{inconsolata}

\usepackage{graphicx}
\usepackage[inline]{enumitem}
\newcommand{\xhdr}[1]{\noindent {\bf #1.} }

\usepackage{booktabs,longtable,tabularx,siunitx,adjustbox,colortbl,xcolor,pdflscape}
\usepackage{caption}
\definecolor{eYouTube}{HTML}{FCEF68}
\definecolor{eCombined}{HTML}{FCB3EB}
\definecolor{eArXiv}{HTML}{FF9824}
\definecolor{eNews}{HTML}{E2AEF4}
\definecolor{dYou}{HTML}{F2FBF4}
\definecolor{dNews}{HTML}{F4F8FF}
\definecolor{dArx}{HTML}{FFF8EE}
\definecolor{dComb}{HTML}{F2F9FF}

\usepackage{siunitx}
\newcommand{\domcell}[1]{%
  \begingroup
  \def\dc{#1}%
  \ifx\dc\youtube\fi
  \endgroup
}

\newcommand{\cnote}[1]{}
\title{Beyond Tokens: Concept-Level Training Objectives for LLMs}

\author{
 \textbf{Laya Iyer},
 \textbf{Pranav Somani},
 \textbf{Alice Guo},
 \textbf{Dan Jurafsky},
 \textbf{Chen Shani}
\\
\{laya, pxsomani, azguo, jurafsky, cshani\} @stanford.edu
\\
 Stanford University
}

\begin{document}
\maketitle
\begin{abstract}
The next-token prediction (NTP) objective has been foundational in the development of modern large language models (LLMs), driving advances in fluency and generalization. 
However, NTP operates at the \textit{token} level, treating deviations from a single reference continuation as errors even when alternative continuations are equally plausible or semantically equivalent (e.g., ``mom'' vs. ``mother'').
As a result, token-level loss can penalize valid abstractions, paraphrases, or conceptually correct reasoning paths, biasing models toward surface form rather than underlying meaning. 
This mismatch between the training signal and semantic correctness motivates learning objectives that operate over higher-level representations.
We propose a shift from token-level to concept-level prediction, where concepts group multiple surface forms of the same idea (e.g., ``mom,'' ``mommy,'' ``mother'' $\rightarrow$ \textit{MOTHER}). 
We introduce various methods for integrating conceptual supervision into LLM training and show that concept-aware models achieve lower perplexity, improved robustness under domain shift, and stronger performance than NTP-based models on diverse NLP benchmarks. 
This suggests \textit{concept-level supervision} as an improved training signal that better aligns LLMs with human semantic abstractions.
\end{abstract}

\section{Introduction}
Large language models (LLMs) have reshaped the landscape of natural language processing, achieving fluency and generalization once thought out of reach. At their core, however, today's LLMs are trained with a surprisingly narrow objective: predicting the next token in a sequence. This has been a powerful proxy for learning language, but it ties models to the surface level of text by rewarding them for producing the right strings, not for understanding the ideas those strings convey. This gap becomes especially pronounced as LLMs are increasingly expected to perform abstraction and reasoning rather than mere continuation.

Humans, by contrast, do not think or communicate in tokens. We reason in \textbf{concepts: semantic units that unify different linguistic expressions under a shared meaning}. For example, ``mom,'' ``mommy,'' and ``mother'' all point to the concept MOTHER. Concepts also stretch beyond literal synonymy: ``father'' may be understood as part of the broader concept PARENT, depending on context. Concepts are flexible, context-sensitive, and hierarchically structured, capturing meaning at a level that tokens cannot \citep{holtzman2021surface, shani2023conceptawarelargelanguagemodels, shani2025tokens, murphy2004big}. 

This gap matters because the expectations placed on LLMs are rapidly shifting. Beyond producing fluent continuations, we now ask them to explain, reason, and think in an abstract manner, tasks that hinge on capturing meaning rather than string similarity. 
In this paper, we explore a different foundation: \textbf{What if models were trained to predict concepts rather than tokens}, by recognizing that multiple forms can stand for the same idea, and to generalize across them? Predicting the next \textit{concept} offers such a shift: instead of optimizing for exact surface matches, models are guided to capture the semantic structures underlying language.

We formalize concepts as clusters of synonymous and contextually interchangeable forms, and integrate them into training as units of supervision. We show that \textbf{predicting the next \textit{concept}, rather than the next token, yields lower NTP perplexity scores, exhibits better robustness to domain shifts, and shows improvements on various NLP benchmarks}. These suggest that concept-aware training can provide a more human-centered foundation for LLM.

\begin{figure*}[t]
  \centering
  \includegraphics[width=\textwidth]{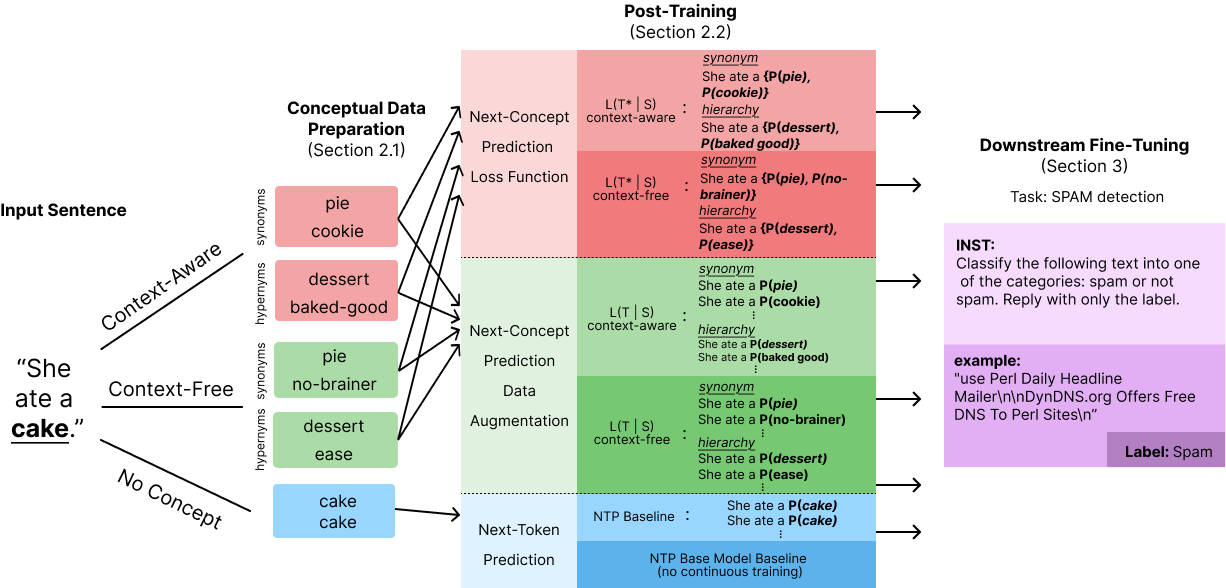} 
  \caption{An outline of our method. We extract context-dependent and independent synonyms and hypernyms and use them to train our NSP and NHP models. We upsample the original data for the NTP baselines. All models are fine-tuned on benchmarks. }
  \label{fig:pipeline}
\end{figure*}

\section{Methods}
We post-trained Llama-3-8B \cite{grattafiori2024llama3herdmodels} using both the standard NTP and our Next-Concept-Prediction (NCP) implementations. To avoid data contamination, we gathered new data that was not used for training the original LLM (as it did not exist then). We now detail the data preparation and training processes (see \hyperref[fig:pipeline]{Figure~\ref*{fig:pipeline}}).\footnote{Data and code: \href{https://github.com/layaiyer1/concept-aware-training}{https://github.com/layaiyer1/concept-aware-training}.}

\subsection{Conceptual Data Preparation}
\label{sec:data_prep}
To avoid training Llama-3-8B on examples from its pretraining corpus, we collected data produced after its public release date (April 18, 2024). Our dataset draws from three distinct sources: YouTube comments, arXiv abstracts, and New York Times abstracts, chosen to provide diversity across informal, scientific, and journalistic domains. 

\subsubsection{Concept Resolution} From this corpus, we extracted nouns from each sentence, treating them as core conceptual units, since nouns typically carry substantial semantic content.\footnote{While verbs and adjectives can also be meaningful, we leave them for future research.} 
We operationalize concepts as interchangeable lexical realizations of a shared concept. Thus, for each noun, we extract two levels of resolution: (1) \textbf{Synonym}, and (2) \textbf{Hypernym}, which is an abstraction of the original noun. Meaning, if the original noun was ``cake,'' synonyms would include ``pie'' and ``cookie,'' whereas the hypernyms would include ``dessert'' and ``baked goods.'' 

\subsubsection{Concept Extraction} As illustrated in \hyperref[fig:pipeline]{Figure 1}, we generated conceptual training data using three complementary methods: (1) \textbf{Context-Free} extracts context-independent, dictionary-based synonyms and hypernyms from WordNet, (2) \textbf{Context-Aware}, which extracts contextual synonyms and hypernyms by prompting Llama-3-8B-Instruct with the full sentence and the target noun to produce contextually appropriate alternatives (\hyperref[sec:contextsynprompt]{Appendix B}), and (3) \textbf{No Concept}, which inflates the data reusing the original word in the sentence, multiple times (to train NTP baselines with matched number of repeated datapoints).\footnote{These methods are imperfect; see 
\hyperref[sec:limitations] {Limitations Section}.}


\subsection{Post-Training}
We post-trained on each dataset split $\in \{$YouTube, {arXiv}, New York Times$\}$ as well as on a combined dataset, downsampling where necessary for consistency across splits. This allows us to assess whether data variation across domains leads to different levels of conceptual awareness. All NCP variants and NTP baselines were post-trained on the same underlying datapoints, differing only in the target nouns used (depending on the concept handling or NTP setup), and on the same number of datapoints (approximately 8K-1K-1K sentences for the train-validation-test split; see Table~\ref{tab:dataset_sizes} for full dataset sizes per variant).

\subsubsection{Next Token Prediction (NTP) Baselines}
We used two NTP-based baselines:

\xhdr{\color{niceblue}Base Model} Used without any post-training to evaluate the model's baseline performance.

\xhdr{\color{niceblue}NTP Baseline} Post-trained using the standard NTP on the same datapoints as the NCT models:
\[
L(T \mid S) = \log \left( p(T \mid S, \Theta) \right)
\]
Where $T$ is the target token, $S$ is the input sentence, and $\Theta$ is model's parameters.

This ensures that the model is exposed to the same datapoints and training volume, while preventing it from learning concept-level signals.



\subsubsection{Next Concept Prediction (NCP) Models}
To shift training from the token level to the concept level, we enriched our data with conceptual signals (see \hyperref[sec:data_prep]{Section~\ref{sec:data_prep}}). Specifically, each target noun $T$ in the corpus was paired with a set of synonym/hypernym nouns $T^*$, extracted either with or without contextual information. Using these conceptual annotations, we implemented two concept-aware training procedures:

\xhdr{\color{nicegreen}NCP Data Augmentation} We augmented the training data using the extracted synonyms and hypernyms. For each sentence with $n$ possible noun completions, we created $n$ training instances, each targeting a different conceptually equivalent noun. The model was then trained using the standard NTP objective. By rewarding these lexical variations, we effectively flattened the probability distribution over next-token predictions, reducing the model's bias toward any single lexical choice.\footnote{We note that data augmentation using synonyms has been explored before by \citet{jungiewicz2019towards, kobayashi2018contextual, levine-etal-2020-sensebert}.}

\xhdr{\color{nicered}NCP Loss Function} A more more direct approach is to modify the NTP loss function itself: Let $T$ debite the original target noun, and the set of conceptually equivalent completions be $T^*$. The objective becomes predicting any completion in $T^*$, conditioned on the input sentence $S$ and the model parameters $\Theta$:
\[
L(T^* \mid S) = \frac{1}{|T^*|} \sum_{n=1}^{N} \log \left( p(t_n \in T^* \mid S, \Theta) \right)
\]

\hyperref[tab:ncp_models]{Table~\ref{tab:ncp_models}} presents all the variants trained using the two NCP training paradigms ({\color{nicegreen}{Data Augmentation}} and {\color{nicered}{Loss Function}}) and two levels of concept resolution (synonyms and hypernyms). 

\begin{table}[t]
\centering
\footnotesize
\begin{tabular}{lll}
\toprule
\textbf{Variant} & \textbf{Model} & \textbf{Concept Method} \\
\midrule
\textbf{NCP Loss} 
& {\color{nicered}NSP Context-Aware} 
& Syn.; LLM \\
& {\color{nicered}NSP Context-Free} 
& Syn.; WordNet \\
& {\color{nicered}NHP Context-Aware} 
& Hyp.; LLM \\
& {\color{nicered}NHP Context-Free} 
& Hyp.; WordNet \\
\midrule
\multirow{4}{*}{\makecell{\textbf{NCP Data}\\\textbf{Augmentation}}}
& {\color{nicegreen}NSP Context-Aware} 
& Syn.; LLM \\
& {\color{nicegreen}NSP Context-Free} 
& Syn.; WordNet \\
& {\color{nicegreen}NHP Context-Aware} 
& Hyp.; LLM \\
& {\color{nicegreen}NHP Context-Free} 
& Hyp.; WordNet \\
\midrule
\multirow{3}{*}{\makecell{\textbf{NTP}\\\textbf{Baselines}}} 
& {\color{niceblue}NTP Synonym} 
& Syn. \\
& {\color{niceblue}NTP Hypernym} 
& Hyp. \\
& {\color{niceblue}Base Model} 
& -- \\
\bottomrule
\end{tabular}
\caption{NCP variants and NTP baselines by post-training paradigm, concept level, and concept source. Syn. = synonym, Hyp. = hypernym. Each variant was post-trained four times, once on each dataset $\in \{$YouTube comments, arXiv abstracts, New York Times abstracts, and a combination of the three domains$\}$.}
\label{tab:ncp_models}
\end{table}

\section{Benchmarks for Fine-Tuning}
\label{sec:data}

After post-training the {\color{niceblue}{NTP baseline}} and all NCP variants ({\color{nicegreen}{Data Augmentation}} and {\color{nicered}{Loss Function}} and the two levels of concept resolution), we fine-tuned them on seven diverse benchmarks (parameters and implementation details in \hyperref[app:ft_imp]{Appendix \ref{app:ft_imp}}):

\xhdr{SNLI} \cite{bowman2015snli} Stanford Natural Language Inference evaluates a model's ability to determine entailment, contradiction, or neutrality between a premise and a hypothesis.

\xhdr{GLUE} \cite{wang2018glue} GLUE aggregates several tasks, such as sentiment analysis, paraphrase detection, and linguistic acceptability, making it a robust testbed for general NLU capabilities.

\xhdr{Empathetic dialogs} \cite{rashkin2019empathic} contains thousands of short conversations grounded in emotional situations, requiring the model to exhibit nuanced understanding and empathetic reasoning. 

\xhdr{Hate speech} \cite{davidson2017automated} composed of tweets annotated for hate speech and offensive language from \url{hatebase.org} and challenges models to distinguish between harmful and benign content.

\xhdr{Spam} \cite{talby2020spamassassin} includes real-world email messages labeled as spam or ham.


\xhdr{Fake News} \cite{cartinoe5930_politifact_fake_news} is built from PolitiFact fact-checks \cite{politifact_site}. It provides news/claims with fake versus real labels on contemporary U.S. political content.

\xhdr{Logical Fallacy} \cite{jin2022logical} reasoning patterns detection dataset spanning ad hominem, ad populum, circular reasoning, false causality, etc.


\section{Results}

We now compare standard NTP with NCP training for both post-training and fine-tuning procedures.

\subsection{Post-Training}

We computed NTP perplexity scores (standard perplexity) on held-out sets from all four domains (YouTube comments, {arXiv} abstracts, and New York Times abstracts, and all together), without modifying the data or enriching it with concept signal. \hyperref[tab:pretrain_transfer]{Table 2} reports the model with the lowest NTP perplexity for each pair of training-test domains across all NCP variants and NCP baselines.

Despite being trained with a different objective and modified data, concept-based models achieve competitive perplexity on held-out subsets of the original data that contain no concept signal. Notably, one model, \textbf{{\color{nicegreen}NHP Context-Aware Data Augmentation}, achieved the lowest NTP perplexity} on all evaluation datasets (all scores in Table~\ref{tab:ppl_full}).

Moreover, we explore cross-domain transfer abilities using the ratio between the NTP\slash NCP perplexity value of the model trained in the evaluated domain and models trained in all other domains. \hyperref[tab:pretrain_transfer]{Table 2} shows that NCP models are better at cross-domain transfer (full scores and metric details in \hyperref[sec:crossdomain]{Appendix~\ref{sec:crossdomain})}.
Overall, \textbf{NCP models demonstrate superior in-domain and cross-domain perplexity scores compared to NTP baselines}.

\begin{table}[h!]
\centering
\label{tab:pretrain_transfer}
\begin{tabular}{ll|c}
\hline
\hline
\textbf{Train\ \ \ \ \ } & \textbf{Eval} & \textbf{Best Model} \\
\hline
\hline
\multirow{4}{*}{\parbox{1pt}{\textbf{YouTube}}}
    & News & \makecell{\color{nicegreen}NSP Context-\\\color{nicegreen}Free Data Aug.}\\
    \cdashline{2-3}
    & ArXiv & {\color{niceblue}NTP Synonym Baseline}\\
    \cdashline{2-3}
    & Combined & \makecell{\color{nicered}NSP Context-Aware\\\& {\color{nicered}NSP Context-Free}}\\
\hline
\multirow{4}{*}{\textbf{News}} 
    & YouTube &  \makecell{\color{nicegreen}NHP Context-\\\color{nicegreen}Free Data Aug.}\\
    \cdashline{2-3}
    & ArXiv & \makecell{\color{nicegreen}NSP Context-\\\color{nicegreen}Aware Data Aug.}\\
    \cdashline{2-3}
    & Combined & \makecell{\color{nicered}NSP Context-Aware\\\& {\color{nicered}NSP Context-Free}}\\
\hline
\multirow{4}{*}{\textbf{ArXiv}}
    & YouTube &  \makecell{\color{nicegreen}NHP Context-\\\color{nicegreen}Free Data Aug.}\\
    \cdashline{2-3}
    & News & \makecell{\color{nicegreen}NSP Context-\\\color{nicegreen}Free Data Aug.}\\
    \cdashline{2-3}
    & Combined &  \makecell{\color{nicered}NSP Context-Aware\\\& {\color{nicered}NSP Context-Free}}\\
\hline
\multirow{4}{*}{Combined} 
    & YouTube &  \makecell{\color{nicegreen}NHP Context-\\\color{nicegreen}Free Data Aug.}\\
    \cdashline{2-3}
    & News & \makecell{\color{nicered}NHP Context-Free}\\
    \cdashline{2-3}
    & ArXiv &  \makecell{\color{niceblue}NTP Synonym Baseline}
\end{tabular}
\caption{\textbf{[NCP models show superior cross domain robustness.]} For each eval domain, we compute the NTP\slash NCP perplexity score of all models \textit{not} trained on the domain and divide them by the corresponding score of the corresponding model that was trained on the eval domain. This captures the robustness to domain shifts.}
\end{table}


\begin{table*}[h!]
\centering
\small
\caption{\textbf{[Incorporating concept-signal into the training process of LLMs improves performance on various downstream NLP tasks.]} Downstream fine-tuned accuracy scores across seven benchmarks: \textsc{Empathetic Dialogues} (EMO), \textsc{GLUE}, \textsc{Hate Speech} (HATE), \textsc{SNLI}, \textsc{SpamAssassin} (SPAM), \textsc{Fake News} (FAKE), \textsc{Logical Fallacy} (LOG). Best accuracy for each dataset within a domain is in bold. A double horizontal line separates the NCP (both NSP and NHP) models from the NCP baselines. NCP models outperform NTP baselines. Notably, the NHP Context-Free Data Augmentation model is best\slash comparable at four out of the seven benchmarks tested. Interestingly, using the combined dataset for post-training does not yield better results compared to using domain-specific datasets.}
\label{tab:downstream-all}
\resizebox{\textwidth}{!}{
\begin{tabular}{ll|rrrrrrr}
\hline
\hline
\textbf{Variant} & \textbf{Domain} & \textbf{EMO} & \textbf{GLUE} & \textbf{HATE} & \textbf{SNLI} & \textbf{SPAM} & \textbf{FAKE} & \textbf{LOG} \\
\hline
\hline
\multirow{4}{*}{\makecell{\color{nicered}NSP Loss\\\color{nicered}Context-Aware}}
  & ArXiv    & 0.8425 & 0.7698 & 0.8544 & 0.8192 & 0.9828 & 0.6431 & 0.4826 \\
  \cdashline{2-9}
  & News     & 0.8183 & 0.8030 & 0.8859 & 0.3515 & 0.9657 & 0.4458 & 0.5149 \\
  \cdashline{2-9}
  & YouTube  & 0.8504 & 0.8433 & 0.8805 & 0.5578 & 0.9532 & 0.6166 & 0.5274 \\
  \cdashline{2-9}
  & Combined & 0.8493 & 0.3365 & 0.9001 & 0.3515 & 0.9782 & 0.5729 & 0.5199 \\
\hline
\multirow{4}{*}{\makecell{\color{nicered}NSP Loss\\\color{nicered}Context-Free}}
  & ArXiv    & 0.8403 & 0.8423 & 0.8490 & 0.4910 & 0.8970 & 0.5987 & 0.4826 \\
  \cdashline{2-9}
  & News     & 0.8575 & 0.8427 & 0.8936 & 0.8250 &  0.6334 & 0.6505 & 0.5348 \\
  \cdashline{2-9}
  & YouTube  & 0.8513 & 0.7844 & 0.8319 & 0.3330 &  0.9657 & 0.6166 & 0.5274 \\
  \cdashline{2-9}
  & Combined & 0.7514 & 0.5879 & 0.7753 & 0.3515 &  0.9657 & 0.576 & 0.1716 \\
\hline
\multirow{4}{*}{\makecell{\color{nicered}NHP Loss\\\color{nicered}Context-Aware}}
  & ArXiv    & 0.8549 & 0.8363 & 0.7806 & 0.6819 &  0.9828 & 0.4263 & 0.5299 \\
  \cdashline{2-9}
  & News     & 0.7846 & \textbf{0.8549} & 0.7952 & 0.8277 &  0.9688 & 0.5757 & 0.5224 \\
  \cdashline{2-9}
  & YouTube  & 0.8043 & 0.3254 & 0.7799 & 0.7948 &  0.9470 & 0.4844 & 0.4950 \\
  \cdashline{2-9}
  & Combined & 0.8425 & 0.8065 & 0.8855 & 0.7450 &  0.9657 & 0.6037 & 0.5274 \\
\hline
\multirow{4}{*}{\makecell{\color{nicered}NHP Loss\\\color{nicered}Context-Free}}
  & ArXiv    & 0.8566 & 0.8363 & 0.7806 & 0.6819 &  0.9828 & 0.4263 & 0.5299 \\
  \cdashline{2-9}
  & News     & 0.8161 & 0.8166 & 0.9001 & 0.3706 &  0.9688 & 0.5757 & 0.4652 \\
  \cdashline{2-9}
  & YouTube  & 0.7767 & 0.3254 & 0.8406 & 0.7948 &  0.9470 & 0.4844 & 0.4950 \\
  \cdashline{2-9}
  & Combined & 0.7745 & 0.8126 & 0.8963 & 0.8271 &  0.8190 & 0.6630 & 0.5323 \\
\hline
\multirow{4}{*}{\makecell{\color{nicegreen}NSP Context-\\\color{nicegreen}Aware Data Aug.}}
  & ArXiv    & {0.8600} & 0.7637 & 0.8532  & 0.8521 & 0.9813 & 0.6197 & 0.5224 \\
  \cdashline{2-9}
  & News     & 0.8566 & 0.6922 & 0.8486 & 0.8362 &  0.9813 & 0.5566 & 0.0970 \\
  \cdashline{2-9}
  & YouTube  & 0.7762 & 0.7652 & 0.8855 & 0.8287 &  0.9750 & 0.6162 & 0.4801 \\
  \cdashline{2-9}
  & Combined & 0.8037 & 0.8433 & 0.8671 & 0.8240 &  0.9204 & 0.6385 & 0.5124 \\
\hline
\multirow{4}{*}{\makecell{\color{nicegreen}NSP Context-\\\color{nicegreen}Free Data Aug.}}
  & ArXiv    & 0.8571 & 0.5829 & 0.8924 & 0.8070 &  0.9672 & 0.6256 & 0.3831 \\
  \cdashline{2-9}
  & News     & 0.8577 & 0.8030 & 0.7787 & 0.8505 &  0.9875 & 0.5764 & \textbf{0.5473} \\
  \cdashline{2-9}
  & YouTube  & 0.8493 & 0.8081 & 0.8598 & 0.8287 &  0.9828 & 0.5858 & 0.5149 \\
  \cdashline{2-9}
  & Combined & 0.7925 & \textbf{0.8574} & 0.8909 & 0.8542 &  0.9782 & 0.5679 & 0.5299 \\
\hline
\multirow{4}{*}{\makecell{\color{nicegreen}NHP Context-\\\color{nicegreen}Aware Data Aug.}}
  & ArXiv    & 0.8093 & 0.4761 & 0.8771 & 0.8287 &  0.9844 & \textbf{0.7176} & 0.4851 \\
  \cdashline{2-9}
  & News     & 0.8397 & 0.8287 & 0.7983 & 0.7004 &  0.9797 & 0.5998 & 0.5174 \\
  \cdashline{2-9}
  & YouTube  & 0.7902 & 0.8363 & 0.8970 & 0.8457 &  0.9064 & 0.6092 & 0.5050 \\
  \cdashline{2-9}
  & Combined & 0.8110 & 0.8111 & 0.8759 & 0.8547 &  0.9189 & 0.6505 & 0.4776 \\
\hline
\multirow{4}{*}{\makecell{\color{nicegreen}NHP Context-\\\color{nicegreen}Free Data Aug.}}
  & ArXiv    & \textbf{0.8673} & 0.6776 & \textbf{0.9051} & 0.7879 &  \textbf{0.9891} & 0.6498 & 0.5149 \\
  \cdashline{2-9}
  & News     & 0.8414 & 0.7662 & 0.8986 & 0.8187 &  0.9766 & 0.5866 & 0.5299 \\
  \cdashline{2-9}
  & YouTube  & 0.8301 & 0.5587 & 0.8940 & \textbf{1} &  0.9844 & 0.5784 & 0.5124 \\
  \cdashline{2-9}
  & Combined & 0.7913 & 0.8282 & 0.8990 & 0.8505 &  0.9485 & 0.5776 & 0.4851 \\
\hline
\hline
\multirow{4}{*}{\makecell{\color{niceblue}NTP Synonym\\\color{niceblue}Baseline Fine-Tuned}}
  & ArXiv    & 0.8313 & 0.7526 & {0.9043} & 0.8388 & 0.8721 & 0.6229 & 0.4876 \\
  \cdashline{2-9}
  & News     & 0.7548 & 0.3496 & 0.8302 & 0.8245 & 0.9111 & 0.6950 & 0.5149 \\
  \cdashline{2-9}
  & YouTube  & 0.8155 & 0.4987 & 0.8798 & 0.8086 & 0.9782 & 0.5761 & 0.4478 \\
  \cdashline{2-9}
  & Combined & 0.8380 & 0.7395 & 0.8509 & 0.8473 & 0.9189 & 0.6076 & 0.5274 \\
\hline
\multirow{4}{*}{\makecell{\color{niceblue}NTP Hypernym\\\color{niceblue}Baseline Fine-Tuned}}
  & ArXiv    & 0.8476 & 0.7149 & \textbf{0.9051} & 0.8388 & 0.9750 & 0.7129 & 0.5174 \\
  \cdashline{2-9}
  & News     & 0.8588 & 0.8262 & 0.8552 & 0.7550 & 0.9797 & 0.5608 & 0.5149 \\
  \cdashline{2-9}
  & YouTube  & 0.8262 & 0.8343 & 0.8986 & 0.8150 & 0.9813 & 0.4395 & 0.5100 \\
  \cdashline{2-9}
  & Combined & 0.7762 & 0.6247 & 0.8944 & 0.8722 & 0.9610 & 0.6392 & 0.4925 \\
\hline
\makecell{\color{niceblue}Base Model\\\color{niceblue}Fine-Tuned} & - & 0.7852 & 0.3365 & 0.8083 & 0.7996 & 0.9813 & 0.6264 & 0.49 \\
\hline
\makecell{\color{niceblue}Base Model} & - & 0.5681 & 0.3204 & 0.7933 & 0.3144 & 0.6505 & 0.4424 & 0.0071 \\
\hline
\end{tabular}}
\end{table*}

To illustrate the qualitative difference between NTP and NCP, consider the following sentence from our data: ``This word has appeared in 53 .''
The NTP's top five predictions are different variations of `articles' (singular versus plural, with and without capitalization and spaces). In contrast, the NCP models distribute the probability mass across semantically related completions: `searches,' `articles,' `episodes,' and `cases,' reflecting a broader conceptual understanding. This highlights that \textbf{while NTP rewards reproducing surface strings, NCP encourages models to capture underlying semantic relationships}, producing outputs that are more meaning-equivalent and less lexically rigid.

\subsection{Downstream Benchmark Fine-Tuning}

\hyperref[tab:downstream-all]{Table 3} reports the accuracy scores of all NCP models and NTP baselines after fine-tuning on the seven benchmarks presented in Section \ref{sec:data}. All models and baselines perform better than the non-fine-tuned variant, as expected. \textbf{Across all seven NLP benchmarks tested, NCP models were consistently better than all NTP baselines}, highlighting the potential of concepts for LLM training.    



\section{Conclusions \& Future Work}

We rethink the standard NTP approach by incorporating more human-inspired supervision signals. We introduce NCP, which unifies synonymous forms into shared semantic units (at two levels of resolution and two training paradigms), enabling models to capture meaning beyond surface text. \textbf{NCP has lower NTP perplexity, is more robust to domain shifts, and exceeds NTP performance, marking it as a promising foundation for LLM training}. Moreover, NCT is flexible, supporting both pre- and post-training applications.

Future work includes NCP pre-training, hierarchical concept representations, and multilingual extensions. We view NTP as one of many possible training signals, whereas \textbf{NCP opens the door to foundations that are not only statistically effective but also more aligned with human cognition}.

\newpage

\newpage
\section{Limitations}
\label{sec:limitations}
While our findings highlight the promise of concept-aware training, several limitations remain.  
First, we explore only one paradigm to incorporate concept supervision. Other formulations, such as hierarchical concepts, cross-lingual mappings, or integration with generative objectives, may provide richer signals.  

Second, our evaluation is limited to fine-tuning on classification tasks. These benchmarks already achieve high baseline accuracy, leaving little room to demonstrate the full potential of concept-level prediction. Extending evaluation to tasks that require more abstraction, such as generation, reasoning, or transfer learning, would offer a clearer picture of its benefits. Broader evaluations and larger-scale experiments are essential to fully establish its effectiveness.  

Third, our approach to extracting concept signals is imperfect. The context-aware method relies on LLMs, which may introduce or amplify existing biases and inconsistencies in their understanding of concepts. The context-free method neglects the crucial role of context in shaping meaning. More robust methods are needed to induce concept representations. 

\section{Ethical Considerations}

In terms of the potential risks of our work, we realize that concepts could lead to the risk of overgeneralizing, overextending concept boundaries, and amplifying spurious associations or stereotypes. Care should be taken when defining concept clusters, especially for sensitive or demographic-related content, to avoid reinforcing biases present in the training data.

We also note that, similar to all NTP LLMs, NCP might lead to hallucinations and other types of undesired model behaviors. These were not explored in this work, and thus we recommend practitioners, as usual, to validate their artifacts before releasing them to the public.

Finally, the introduction of concept-level reasoning may shift the interpretability of model outputs: while grouping tokens into concepts can improve semantic coherence, it may obscure the model’s reasoning at the token level, potentially making errors harder to detect. We encourage transparency in reporting both concept definitions and model behaviors to support responsible use.

Disclosure: LLMs were used to refine the text and design the table. 

\bibliography{custom}

@article{kobayashi2018contextual,
  title={Contextual augmentation: Data augmentation by words with paradigmatic relations},
  author={Kobayashi, Sosuke},
  journal={arXiv preprint arXiv:1805.06201},
  year={2018}
}

@article{jungiewicz2019towards,
  title={Towards textual data augmentation for neural networks: synonyms and maximum loss},
  author={Jungiewicz, Micha{\l} and Smywi{\'n}ski-Pohl, Aleksander},
  journal={Computer Science},
  volume={20},
  year={2019}
}

@article{shani2025tokens,
  title={From tokens to thoughts: How LLMs and humans trade compression for meaning},
  author={Shani, Chen and Soffer, Liron and Jurafsky, Dan and LeCun, Yann and Shwartz-Ziv, Ravid},
  journal={arXiv preprint arXiv:2505.17117},
  year={2025}
}

@book{murphy2004big,
  title={The big book of concepts},
  author={Murphy, Gregory},
  year={2004},
  publisher={MIT press}
}

@inproceedings{levine-etal-2020-sensebert,
    title = "{S}ense{BERT}: Driving Some Sense into {BERT}",
    author = "Levine, Yoav  and
      Lenz, Barak  and
      Dagan, Or  and
      Ram, Ori  and
      Padnos, Dan  and
      Sharir, Or  and
      Shalev-Shwartz, Shai  and
      Shashua, Amnon  and
      Shoham, Yoav",
    editor = "Jurafsky, Dan  and
      Chai, Joyce  and
      Schluter, Natalie  and
      Tetreault, Joel",
    booktitle = "Proceedings of the 58th Annual Meeting of the Association for Computational Linguistics",
    month = jul,
    year = "2020",
    address = "Online",
    publisher = "Association for Computational Linguistics",
    url = "https://aclanthology.org/2020.acl-main.423/",
    doi = "10.18653/v1/2020.acl-main.423",
    pages = "4656--4667"
}

@article{holtzman2021surface,
  title={Surface form competition: Why the highest probability answer isn't always right},
  author={Holtzman, Ari and West, Peter and Shwartz, Vered and Choi, Yejin and Zettlemoyer, Luke},
  journal={arXiv preprint arXiv:2104.08315},
  year={2021}
}

@misc{hu2021loralowrankadaptationlarge,
      title={LoRA: Low-Rank Adaptation of Large Language Models}, 
      author={Edward J. Hu and Yelong Shen and Phillip Wallis and Zeyuan Allen-Zhu and Yuanzhi Li and Shean Wang and Lu Wang and Weizhu Chen},
      year={2021},
      eprint={2106.09685},
      archivePrefix={arXiv},
      primaryClass={cs.CL},
      url={https://arxiv.org/abs/2106.09685}, 
}

@misc{shani2023conceptawarelargelanguagemodels,
      title={Towards Concept-Aware Large Language Models}, 
      author={Chen Shani and Jilles Vreeken and Dafna Shahaf},
      year={2023},
      eprint={2311.01866},
      archivePrefix={arXiv},
      primaryClass={cs.CL},
      url={https://arxiv.org/abs/2311.01866}, 
}

@inproceedings{bowman2015snli,
  title={A large annotated corpus for learning natural language inference},
  author={Bowman, Samuel R and Angeli, Gabor and Potts, Christopher and Manning, Christopher D},
  booktitle={Proceedings of the 2015 Conference on Empirical Methods in Natural Language Processing},
  pages={632--642},
  year={2015},
  organization={ACL}
}

@inproceedings{wang2018glue,
  title={GLUE: A multi-task benchmark and analysis platform for natural language understanding},
  author={Wang, Alex and Singh, Amanpreet and Michael, Julian and Hill, Felix and Levy, Omer and Bowman, Samuel R},
  booktitle={Proceedings of the 2018 EMNLP Workshop BlackboxNLP},
  pages={353--355},
  year={2018},
  organization={ACL}
}

@inproceedings{rashkin2019empathic,
  title={Towards empathetic open-domain conversation models: A new benchmark and dataset},
  author={Rashkin, Hannah and Smith, Eric Michael and Li, Margaret and Boureau, Y-Lan},
  booktitle={Proceedings of the 57th Annual Meeting of the Association for Computational Linguistics},
  pages={5370--5381},
  year={2019},
  organization={ACL}
}

@inproceedings{davidson2017automated,
  title={Automated Hate Speech Detection and the Problem of Offensive Language},
  author={Davidson, Thomas and Warmsley, Dana and Macy, Michael and Weber, Ingmar},
  booktitle={Proceedings of ICWSM},
  year={2017}
}

@misc{talby2020spamassassin,
  title        = {SpamAssassin Public Corpus Dataset},
  author       = {Talby, David},
  year         = {2020},
  howpublished = {\url{https://huggingface.co/datasets/talby/spamassassin}}
}

@dataset{mafi2023suicidal,
  author = {Mafi, Md Mafiul Hasan Matin and Alam, Md. Sabbir},
  title = {Suicidal Ideation Detection Reddit Dataset},
  year = {2023},
  publisher = {Mendeley Data},
  version = {V2},
  doi = {10.17632/z8s6w86tr3.2},
  url = {https://doi.org/10.17632/z8s6w86tr3.2}
}

@misc{grattafiori2024llama3herdmodels,
      title={The Llama 3 Herd of Models}, 
      author={Aaron Grattafiori and Abhimanyu Dubey and Abhinav Jauhri and Abhinav Pandey and Abhishek Kadian and Ahmad Al-Dahle and Aiesha Letman and Akhil Mathur and Alan Schelten and Alex Vaughan and Amy Yang and Angela Fan and Anirudh Goyal and Anthony Hartshorn and Aobo Yang and Archi Mitra and Archie Sravankumar and Artem Korenev and Arthur Hinsvark and Arun Rao and Aston Zhang and Aurelien Rodriguez and Austen Gregerson and Ava Spataru and Baptiste Roziere and Bethany Biron and Binh Tang and Bobbie Chern and Charlotte Caucheteux and Chaya Nayak and Chloe Bi and Chris Marra and Chris McConnell and Christian Keller and Christophe Touret and Chunyang Wu and Corinne Wong and Cristian Canton Ferrer and Cyrus Nikolaidis and Damien Allonsius and Daniel Song and Danielle Pintz and Danny Livshits and Danny Wyatt and David Esiobu and Dhruv Choudhary and Dhruv Mahajan and Diego Garcia-Olano and Diego Perino and Dieuwke Hupkes and Egor Lakomkin and Ehab AlBadawy and Elina Lobanova and Emily Dinan and Eric Michael Smith and Filip Radenovic and Francisco Guzmán and Frank Zhang and Gabriel Synnaeve and Gabrielle Lee and Georgia Lewis Anderson and Govind Thattai and Graeme Nail and Gregoire Mialon and Guan Pang and Guillem Cucurell and Hailey Nguyen and Hannah Korevaar and Hu Xu and Hugo Touvron and Iliyan Zarov and Imanol Arrieta Ibarra and Isabel Kloumann and Ishan Misra and Ivan Evtimov and Jack Zhang and Jade Copet and Jaewon Lee and Jan Geffert and Jana Vranes and Jason Park and Jay Mahadeokar and Jeet Shah and Jelmer van der Linde and Jennifer Billock and Jenny Hong and Jenya Lee and Jeremy Fu and Jianfeng Chi and Jianyu Huang and Jiawen Liu and Jie Wang and Jiecao Yu and Joanna Bitton and Joe Spisak and Jongsoo Park and Joseph Rocca and Joshua Johnstun and Joshua Saxe and Junteng Jia and Kalyan Vasuden Alwala and Karthik Prasad and Kartikeya Upasani and Kate Plawiak and Ke Li and Kenneth Heafield and Kevin Stone and Khalid El-Arini and Krithika Iyer and Kshitiz Malik and Kuenley Chiu and Kunal Bhalla and Kushal Lakhotia and Lauren Rantala-Yeary and Laurens van der Maaten and Lawrence Chen and Liang Tan and Liz Jenkins and Louis Martin and Lovish Madaan and Lubo Malo and Lukas Blecher and Lukas Landzaat and Luke de Oliveira and Madeline Muzzi and Mahesh Pasupuleti and Mannat Singh and Manohar Paluri and Marcin Kardas and Maria Tsimpoukelli and Mathew Oldham and Mathieu Rita and Maya Pavlova and Melanie Kambadur and Mike Lewis and Min Si and Mitesh Kumar Singh and Mona Hassan and Naman Goyal and Narjes Torabi and Nikolay Bashlykov and Nikolay Bogoychev and Niladri Chatterji and Ning Zhang and Olivier Duchenne and Onur Çelebi and Patrick Alrassy and Pengchuan Zhang and Pengwei Li and Petar Vasic and Peter Weng and Prajjwal Bhargava and Pratik Dubal and Praveen Krishnan and Punit Singh Koura and Puxin Xu and Qing He and Qingxiao Dong and Ragavan Srinivasan and Raj Ganapathy and Ramon Calderer and Ricardo Silveira Cabral and Robert Stojnic and Roberta Raileanu and Rohan Maheswari and Rohit Girdhar and Rohit Patel and Romain Sauvestre and Ronnie Polidoro and Roshan Sumbaly and Ross Taylor and Ruan Silva and Rui Hou and Rui Wang and Saghar Hosseini and Sahana Chennabasappa and Sanjay Singh and Sean Bell and Seohyun Sonia Kim and Sergey Edunov and Shaoliang Nie and Sharan Narang and Sharath Raparthy and Sheng Shen and Shengye Wan and Shruti Bhosale and Shun Zhang and Simon Vandenhende and Soumya Batra and Spencer Whitman and Sten Sootla and Stephane Collot and Suchin Gururangan and Sydney Borodinsky and Tamar Herman and Tara Fowler and Tarek Sheasha and Thomas Georgiou and Thomas Scialom and Tobias Speckbacher and Todor Mihaylov and Tong Xiao and Ujjwal Karn and Vedanuj Goswami and Vibhor Gupta and Vignesh Ramanathan and Viktor Kerkez and Vincent Gonguet and Virginie Do and Vish Vogeti and Vítor Albiero and Vladan Petrovic and Weiwei Chu and Wenhan Xiong and Wenyin Fu and Whitney Meers and Xavier Martinet and Xiaodong Wang and Xiaofang Wang and Xiaoqing Ellen Tan and Xide Xia and Xinfeng Xie and Xuchao Jia and Xuewei Wang and Yaelle Goldschlag and Yashesh Gaur and Yasmine Babaei and Yi Wen and Yiwen Song and Yuchen Zhang and Yue Li and Yuning Mao and Zacharie Delpierre Coudert and Zheng Yan and Zhengxing Chen and Zoe Papakipos and Aaditya Singh and Aayushi Srivastava and Abha Jain and Adam Kelsey and Adam Shajnfeld and Adithya Gangidi and Adolfo Victoria and Ahuva Goldstand and Ajay Menon and Ajay Sharma and Alex Boesenberg and Alexei Baevski and Allie Feinstein and Amanda Kallet and Amit Sangani and Amos Teo and Anam Yunus and Andrei Lupu and Andres Alvarado and Andrew Caples and Andrew Gu and Andrew Ho and Andrew Poulton and Andrew Ryan and Ankit Ramchandani and Annie Dong and Annie Franco and Anuj Goyal and Aparajita Saraf and Arkabandhu Chowdhury and Ashley Gabriel and Ashwin Bharambe and Assaf Eisenman and Azadeh Yazdan and Beau James and Ben Maurer and Benjamin Leonhardi and Bernie Huang and Beth Loyd and Beto De Paola and Bhargavi Paranjape and Bing Liu and Bo Wu and Boyu Ni and Braden Hancock and Bram Wasti and Brandon Spence and Brani Stojkovic and Brian Gamido and Britt Montalvo and Carl Parker and Carly Burton and Catalina Mejia and Ce Liu and Changhan Wang and Changkyu Kim and Chao Zhou and Chester Hu and Ching-Hsiang Chu and Chris Cai and Chris Tindal and Christoph Feichtenhofer and Cynthia Gao and Damon Civin and Dana Beaty and Daniel Kreymer and Daniel Li and David Adkins and David Xu and Davide Testuggine and Delia David and Devi Parikh and Diana Liskovich and Didem Foss and Dingkang Wang and Duc Le and Dustin Holland and Edward Dowling and Eissa Jamil and Elaine Montgomery and Eleonora Presani and Emily Hahn and Emily Wood and Eric-Tuan Le and Erik Brinkman and Esteban Arcaute and Evan Dunbar and Evan Smothers and Fei Sun and Felix Kreuk and Feng Tian and Filippos Kokkinos and Firat Ozgenel and Francesco Caggioni and Frank Kanayet and Frank Seide and Gabriela Medina Florez and Gabriella Schwarz and Gada Badeer and Georgia Swee and Gil Halpern and Grant Herman and Grigory Sizov and Guangyi and Zhang and Guna Lakshminarayanan and Hakan Inan and Hamid Shojanazeri and Han Zou and Hannah Wang and Hanwen Zha and Haroun Habeeb and Harrison Rudolph and Helen Suk and Henry Aspegren and Hunter Goldman and Hongyuan Zhan and Ibrahim Damlaj and Igor Molybog and Igor Tufanov and Ilias Leontiadis and Irina-Elena Veliche and Itai Gat and Jake Weissman and James Geboski and James Kohli and Janice Lam and Japhet Asher and Jean-Baptiste Gaya and Jeff Marcus and Jeff Tang and Jennifer Chan and Jenny Zhen and Jeremy Reizenstein and Jeremy Teboul and Jessica Zhong and Jian Jin and Jingyi Yang and Joe Cummings and Jon Carvill and Jon Shepard and Jonathan McPhie and Jonathan Torres and Josh Ginsburg and Junjie Wang and Kai Wu and Kam Hou U and Karan Saxena and Kartikay Khandelwal and Katayoun Zand and Kathy Matosich and Kaushik Veeraraghavan and Kelly Michelena and Keqian Li and Kiran Jagadeesh and Kun Huang and Kunal Chawla and Kyle Huang and Lailin Chen and Lakshya Garg and Lavender A and Leandro Silva and Lee Bell and Lei Zhang and Liangpeng Guo and Licheng Yu and Liron Moshkovich and Luca Wehrstedt and Madian Khabsa and Manav Avalani and Manish Bhatt and Martynas Mankus and Matan Hasson and Matthew Lennie and Matthias Reso and Maxim Groshev and Maxim Naumov and Maya Lathi and Meghan Keneally and Miao Liu and Michael L. Seltzer and Michal Valko and Michelle Restrepo and Mihir Patel and Mik Vyatskov and Mikayel Samvelyan and Mike Clark and Mike Macey and Mike Wang and Miquel Jubert Hermoso and Mo Metanat and Mohammad Rastegari and Munish Bansal and Nandhini Santhanam and Natascha Parks and Natasha White and Navyata Bawa and Nayan Singhal and Nick Egebo and Nicolas Usunier and Nikhil Mehta and Nikolay Pavlovich Laptev and Ning Dong and Norman Cheng and Oleg Chernoguz and Olivia Hart and Omkar Salpekar and Ozlem Kalinli and Parkin Kent and Parth Parekh and Paul Saab and Pavan Balaji and Pedro Rittner and Philip Bontrager and Pierre Roux and Piotr Dollar and Polina Zvyagina and Prashant Ratanchandani and Pritish Yuvraj and Qian Liang and Rachad Alao and Rachel Rodriguez and Rafi Ayub and Raghotham Murthy and Raghu Nayani and Rahul Mitra and Rangaprabhu Parthasarathy and Raymond Li and Rebekkah Hogan and Robin Battey and Rocky Wang and Russ Howes and Ruty Rinott and Sachin Mehta and Sachin Siby and Sai Jayesh Bondu and Samyak Datta and Sara Chugh and Sara Hunt and Sargun Dhillon and Sasha Sidorov and Satadru Pan and Saurabh Mahajan and Saurabh Verma and Seiji Yamamoto and Sharadh Ramaswamy and Shaun Lindsay and Shaun Lindsay and Sheng Feng and Shenghao Lin and Shengxin Cindy Zha and Shishir Patil and Shiva Shankar and Shuqiang Zhang and Shuqiang Zhang and Sinong Wang and Sneha Agarwal and Soji Sajuyigbe and Soumith Chintala and Stephanie Max and Stephen Chen and Steve Kehoe and Steve Satterfield and Sudarshan Govindaprasad and Sumit Gupta and Summer Deng and Sungmin Cho and Sunny Virk and Suraj Subramanian and Sy Choudhury and Sydney Goldman and Tal Remez and Tamar Glaser and Tamara Best and Thilo Koehler and Thomas Robinson and Tianhe Li and Tianjun Zhang and Tim Matthews and Timothy Chou and Tzook Shaked and Varun Vontimitta and Victoria Ajayi and Victoria Montanez and Vijai Mohan and Vinay Satish Kumar and Vishal Mangla and Vlad Ionescu and Vlad Poenaru and Vlad Tiberiu Mihailescu and Vladimir Ivanov and Wei Li and Wenchen Wang and Wenwen Jiang and Wes Bouaziz and Will Constable and Xiaocheng Tang and Xiaojian Wu and Xiaolan Wang and Xilun Wu and Xinbo Gao and Yaniv Kleinman and Yanjun Chen and Ye Hu and Ye Jia and Ye Qi and Yenda Li and Yilin Zhang and Ying Zhang and Yossi Adi and Youngjin Nam and Yu and Wang and Yu Zhao and Yuchen Hao and Yundi Qian and Yunlu Li and Yuzi He and Zach Rait and Zachary DeVito and Zef Rosnbrick and Zhaoduo Wen and Zhenyu Yang and Zhiwei Zhao and Zhiyu Ma},
      year={2024},
      eprint={2407.21783},
      archivePrefix={arXiv},
      primaryClass={cs.AI},
      url={https://arxiv.org/abs/2407.21783}, 
}

@article{jin2022logical,
  title={Logical Fallacy Detection},
  author={Jin, Zhijing and Lalwani, Abhinav and Vaidhya, Tejas and Shen, Xiaoyu and Ding, Yiwen and Lyu, Zhiheng and Sachan, Mrinmaya and Mihalcea, Rada and Sch{\"o}lkopf, Bernhard},
  journal={arXiv preprint arXiv:2202.13758},
  year={2022}
}

@inproceedings{zhang2015character,
  title={Character-level Convolutional Networks for Text Classification},
  author={Zhang, Xiang and Zhao, Junbo and LeCun, Yann},
  booktitle={NeurIPS},
  year={2015}
}

@misc{cartinoe5930_politifact_fake_news,
  author       = {Cartinoe5930},
  title        = {PolitiFact Fake News},
  howpublished = {Hugging Face Datasets},
  year         = {2025},
  url          = {https://huggingface.co/datasets/Cartinoe5930/Politifact_fake_news},
  note         = {Accessed: 2025-10-06}
}

@online{politifact_site,
  author    = {{PolitiFact}},
  title     = {PolitiFact: Fact-checks and Truth-O-Meter},
  year      = {2007--2025},
  url       = {https://www.politifact.com/},
  note      = {Accessed: 2025-10-06}
}

\clearpage
\appendix

\section{Prompt to Obtain Contextual Synonym and Hypernym}
\label{sec:contextsynprompt}

\subsection*{Synonyms}
\paragraph{System Prompt}
\texttt{Answer the question using a comma-separated list and remove any extraneous information. An example output for a sentence will be [item1, item2, item3].  If no synonyms are found, return an empty array. Do not repeat this prompt in your output.}
\paragraph{Message}
Provided a \textbf{sentence} and a \textbf{noun} of interest, the message reads:
\texttt{"Generate contextual synonyms for the word \textbf{noun} in the sentence \textbf{sentence}."}

\subsection*{Hypernyms}
\paragraph{System Prompt}
\texttt{Answer the question using a comma-separated list and remove any extraneous information. An example output for a sentence will be [item1, item2, item3].  If no hypernym are found, return an empty array. Do not repeat this prompt in your output.}
\paragraph{Message}
Provided a \textbf{sentence} and a \textbf{noun} of interest, the message reads:
\texttt{"Generate contextual hypernym for the word \textbf{noun} in the sentence \textbf{sentence}."}



\section{Fine-Tuning Implementation}
\label{app:ft_imp}

For each of the following tasks, we fine-tuned all models using LoRA \cite{hu2021loralowrankadaptationlarge} with parameters r=16 and $\alpha$=16, targeting the attention and feed-forward modules (q\_proj, k\_proj, v\_proj, o\_proj, gate\_proj, up\_proj, down\_proj) for efficient adaptation. Models were trained using 4-bit quantization with the AdamW 8-bit optimizer, a learning rate of 2e-4 with linear scheduling, and gradient accumulation over four steps. Each model was trained for 100 steps with a batch size of 2, employing the Alpaca instruction format for consistent prompt structuring across tasks. Training incorporated validation-based checkpointing every 20 steps to monitor convergence. This resulted in a total of 189 fine-tuned models across 9 downstream tasks, enabling us to systematically evaluate the transferability and robustness of conceptual understanding across domains.
 
 We evaluated each fine-tuned model's ability to classify instances for the task for which it was trained. The evaluation process involved matching each model to its input template and generating predictions using the Alpaca prompt format. We computed match accuracy by comparing lowercased, stripped predictions against ground truth labels. The evaluation focused on accuracy as the primary metric for comparing the concept-aware training paradigm against baseline approaches, with results stored in JSON format, including sample predictions for qualitative analysis. This systematic evaluation enabled direct comparison of how different post-training strategies transferred to downstream classification tasks.

\section{Post-training Dataset Sizes}
\hyperref[tab:dataset_sizes]{Table}~\ref{tab:dataset_sizes} depicts the different dataset sizes for each domain and model variant. These are the same dataset sizes for both the synonym based models, as well as the hierarchy based models.

\begin{table}[H]
\centering
\caption{\textbf{Dataset sizes for each domain and model variant.}}
\label{tab:dataset_sizes}
\begin{tabular}{llrrr}
\hline\hline
\textbf{Domain} & \textbf{Variant} & \textbf{Train} & \textbf{Val} & \textbf{Test (Eval)} \\
\hline\hline
\multirow{3}{*}{ArXiv}
    & Vanilla & 8{,}000 & 1{,}002 & 986 \\
    & Dict    & 8{,}000 & 1{,}002 & 986 \\
    & Context & 8{,}006 & 1{,}002 & 986 \\
\hline
\multirow{3}{*}{News}
    & Vanilla & 8{,}001 & 1{,}007 & 1{,}001 \\
    & Dict    & 8{,}001 & 1{,}007 & 1{,}001 \\
    & Context & 9{,}989 & 1{,}007 & 1{,}001 \\
\hline
\multirow{3}{*}{YouTube}
    & Vanilla & 8{,}000 & 1{,}004 & 964 \\
    & Dict    & 8{,}000 & 1{,}004 & 964 \\
    & Context & 8{,}000 & 1{,}004 & 964 \\
\hline
\multirow{3}{*}{Combined}
    & Vanilla & 8{,}000 & 1{,}002 & 986 \\
    & Dict    & 8{,}000 & 1{,}002 & 986 \\
    & Context & 8{,}004 & 1{,}002 & 986 \\
\hline
\end{tabular}
\end{table}

\section{Multi-Token Completions}
Our loss function supports \emph{multi-token} words and completions. We precompute a map word $\rightarrow$ token IDs for all targets and their completions to avoid repeated tokenization and keep training steps fast/deterministic. Using this dictionary from words (nouns of interest) to their tokenized IDs we replace the entire token span of the target word with the completion’s span (which may be longer or shorter) during training. The model is then evaluated using the completions and the NCP loss function.

\section{Model Size and Budget}
In this paper, we use LLaMA-3-8B as our main model, which is an 8-billion-parameter model. Computational resources and GPUs were provided by the authors' research institute.

\section{Dataset Descriptive Statistics}
The following is additional details about each of our fine-tuning datasets:
\paragraph{SNLI \cite{bowman2015snli}}(\href{https://huggingface.co/datasets/stanfordnlp/snli}{\texttt{stanfordnlp/snli}})
\begin{itemize}
    \item \textbf{Task:} 3-way NLI (entailment/contradiction/neutral).
    \item \textbf{Size/Splits/Labels:} $\sim$570k pairs; train/dev/test; labels: \textit{entailment, contradiction, neutral}.
    \item \textbf{Examples from the dataset:} \\
    (1) \texttt{Premise: A man inspects the uniform of a figure in an East Asian country. Hypothesis: The man is sleeping} \\\textbf{Label:} Contradiction\\\\
    (2) \texttt{Premise: Two men on a roof with snow shovels. Hypothesis: They are clearing snow.} \\\textbf{Label:} Entailment
\end{itemize}

\paragraph{GLUE \cite{wang2018glue}}(\href{https://huggingface.co/datasets/nyu-mll/glue}{\texttt{nyu-mll/glue}})
\begin{itemize}
    \item \textbf{Task:} Aggregated NLU suite (acceptability, sentiment, paraphrase, NLI, STS).
    \item \textbf{Size/Splits/Labels:} 13.2k pairs; train/dev/test; labels: entailment, contradiction, neutral.
    \item \textbf{Examples from the dataset:} \\
    (1) \texttt{Premise: but see but you're going there and you know what you're getting into. Hypothesis: By getting involved, you understand what is in store.} \\\textbf{Label:} Entailment\\\\
    (2) \texttt{Premise: it is in Texas too. Hypothesis: It's not in Texas} \\\textbf{Label:} Contradiction
\end{itemize}

\paragraph{Empathetic Dialogues \cite{rashkin2019empathic}}(\href{https://huggingface.co/datasets/facebook/empathetic_dialogues}{\texttt{facebook/empathetic\_dialogues}})
\begin{itemize}
    \item \textbf{Task:} Emotion-grounded open-domain dialogue.
    \item \textbf{Size/Splits/Labels:} 76.7k/12.0k/10.9k (train/dev/test); emotion in \texttt{context}.
    \item \textbf{Examples from the dataset:} \\
    (1) \texttt{Utterance: I remember going to see the fireworks with my best friend.} \\\textbf{Label:} Sentimental\\\\
    (2) \texttt{Utterance: I finally finished my last exam today!} \\\textbf{Label:} Proud
\end{itemize}

\paragraph{Hate Speech \cite{davidson2017automated}}(\href{https://huggingface.co/datasets/tdavidson/hate_speech_offensive}{\texttt{tdavidson/hate\_speech\_offensive}})
\begin{itemize}
    \item \textbf{Task:} Tweet toxicity (hate\_speech/offensive/neither)
    \item \textbf{Size/Splits/Labels:} train/dev/test; 3 labels.
    \item \textbf{Examples from the dataset:} \\
    (1) \texttt{Input: @user we gotta find this h**.} \\\textbf{Label:} Offensive\\\\
    (2) \texttt{Input: Burritos are trash} \\\textbf{Label:} Neither
    \item \textbf{Content note:} Contains offensive language.
\end{itemize}

\paragraph{Spam Assassin \cite{talby2020spamassassin}}(\href{https://huggingface.co/datasets/talby/spamassassin}{\texttt{talby/spamassassin}})
\begin{itemize}
    \item \textbf{Task:} Spam vs.\ ham email classification.
    \item \textbf{Size/Splits/Labels:} $\sim$21.5k messages; labels: \textit{spam, ham}.
    \item \textbf{Examples from the dataset:} \\
    (1) \texttt{Input: Free trial for~\ldots{}} \\\textbf{Label:} Spam\\\\
    (2) \texttt{Input: Meeting moved to 3pm~\ldots{}} \\\textbf{Label:} Ham
\end{itemize}

\paragraph{Suicidal Ideation (Reddit) \cite{mafi2023suicidal}}(\href{https://data.mendeley.com/datasets/z8s6w86tr3}{Mendeley Data (DOI)})
\begin{itemize}
    \item \textbf{Task:} Spam vs.\ ham email classification.
    \item \textbf{Size/Splits/Labels:} 15{,}477 posts (paper).
    \item \textbf{Examples from the dataset:} \\
    (1) \texttt{Input: “I can’t see a way out~\ldots{} I’m so tired.”} \\\textbf{Label:} Suicidal\\\\
    (2) \texttt{Input: "Having a rough day but trying to stay positive."} \\\textbf{Label:} Non-suicidal
    \item Content note: Sensitive mental-health content.
\end{itemize}

\paragraph{Fake News (PolitiFact-based) \cite{cartinoe5930_politifact_fake_news}}(\href{https://huggingface.co/datasets/Cartinoe5930/Politifact_fake_news}{Cartinoe5930/Politifact\_fake\_news})
\begin{itemize}
    \item \textbf{Task:} Short political claim with fact-check label (e.g.; \textit{true}, \textit{false})
    \item \textbf{Size/Splits/Labels:} train 17.1k, text 4.23k; labels: true or false.
    \item \textbf{Examples from the dataset:} \\
    (1) \texttt{Input: PayPal has reinstated its policy to fine users \$2,500 directly from their accounts if they spread 'misinformation.} \\\textbf{Label:} False\\\\
    (2) \texttt{Input: Kids are resistant to COVID as opposed to older people.} \\\textbf{Label:} True
\end{itemize}

\paragraph{Logical Fallacy \cite{jin2022logical}}(\href{https://huggingface.co/datasets/tasksource/logical-fallacy}{tasksource/logical-fallacy})
\begin{itemize}
    \item \textbf{Task:} Multi-class fallacy detection (e.g.; ad hominem, ad populum, circular reasoning, false causality)
    \item \textbf{Size/Splits/Labels:} train 2.68k, test 500; labels: ad hominem, ad populum, appeal to emotion, circular reasoning, equivocation, fallacy of credibility, fallacy of extension, fallacy of logic, fallacy of relevance, false causality, false dilemma, faulty generalization, intentional.
    \item \textbf{Examples from the dataset: }\\
    (1) \texttt{Input: Don't listen to Senator Bob's opinion. He is a crook, and a spiteful loony man.}  \\\textbf{Label:} ad hominem\\\\
    (2) \texttt{Input: Did your misleading claims result in you getting promoted?} \\\textbf{Label:} intentional
\end{itemize}

\paragraph{Amazon Polarity \cite{zhang2015character}}(\href{https://huggingface.co/datasets/amazon_polarity}{amazon\_polarity})
\begin{itemize}
    \item \textbf{Task:} Binary review sentiment.
    \item \textbf{Size/Splits/Labels:} train 3.6M, test 400k; labels: positive, negative.
    \item \textbf{Examples from the dataset:} \\
    (1) \texttt{Input: A complete waste of time. Typographical errors, poor grammar, and a totally pathetic plot add up to absolutely nothing. I'm embarrassed for this author and very disappointed I actually paid for this book.} \\\textbf{Label:} negative\\\\
    (2) \texttt{Input: got this for my daughter in NC, she is now making prefect bread. Wish she lived closer to make me some} \\\textbf{Label:} positive
\end{itemize}

\section{Cross-Domain Table}
\label{sec:crossdomain}

Post-training perplexity scores using both the standard NTP and our NCP for calculating these perplexity scores. The rightmost column depicts the domain-shift robustness and is calculated as follows: the perplexity score (using the relevant N\_P; NTP for baselines and NCP for all others) trained on a different domain than the evaluation is divided by the perplexity score of the corresponding model that was trained on the domain. This allows us to normalize the perplexity in a way that only preserves robustness to domain shifts. For example, the NTP perplexity score of the NTP baseline trained on \textit{news} and evaluated on \textit{YouTube} is 244.5672. We divide it by the NTP perplexity score of the NTP baseline trained on \textit{YouTube} (184.3536), resulting in 1.32662015. Similar to perplexity scores, lower numbers indicate better robustness to domain shifts.

\pagebreak
{\small
\onecolumn
\begin{longtable}{@{}>{\raggedright\arraybackslash}p{1.2cm}
                  >{\raggedright\arraybackslash}p{3.2cm}
                  >{\raggedright\arraybackslash}p{1.4cm}
                  >{\raggedright\arraybackslash}p{1.25cm}
                  S[table-format=7.4,table-number-alignment=center]
                  S[table-format=8.4,table-number-alignment=center]
                  S[table-format=2.4,table-column-width=2.4cm,table-number-alignment=center]@{}}
\captionsetup{justification=raggedright, singlelinecheck=false}
\caption{Post-training perplexity (PPL) with NTP and concept-resolution objectives (for NCP -- either NHP/NSP). The last column (N\_P/Domain N\_P) shows cross-domain transfer: PPL of a model trained on a source domain and evaluated on a target, normalized by the model trained (and evaluated) on that target with the same objective.}
\label{tab:ppl_full}\\
\toprule
\textbf{Concept} & \textbf{Variant} & \textbf{Evaluated} & \textbf{Domain} & \textbf{NTP PPL} & \textbf{NCP PPL} & \textbf{N\_P/Domain N\_P} \\
\midrule
\endfirsthead
\toprule
\textbf{Concept} & \textbf{Variant} & \textbf{Evaluated} & \textbf{Domain} & \textbf{NTP PPL} & \textbf{NHP/NSP PPL} & \textbf{N\_P/Domain N\_P} \\
\midrule
\endhead
\midrule
\multicolumn{7}{r}{\small (continued)}\\
\midrule
\endfoot
\bottomrule
\endlastfoot

NHP & \textcolor{nicered}{Context-Aware} & \cellcolor{eYouTube}YouTube & \cellcolor{eYouTube}youtube & 333.2447 & 2.7490 & 1 \\
NHP & \textcolor{nicered}{Context-Free} & \cellcolor{eYouTube}YouTube & \cellcolor{eYouTube}youtube & 333.2447 & 2.7490 & 1 \\
NHP & \textcolor{nicegreen}{Context-Aware Data-Aug} & \cellcolor{eYouTube}YouTube & \cellcolor{eYouTube}youtube & 92.2984 & 280477.1748 & 1 \\
NHP & \textcolor{nicegreen}{Context-Free Data-Aug} & \cellcolor{eYouTube}YouTube & \cellcolor{eYouTube}youtube & 89.6938 & 350333.1050 & 1 \\
NTP & \textcolor{niceblue}{Hypernyms Baseline} & \cellcolor{eYouTube}YouTube & \cellcolor{eYouTube}youtube & 85.1368 & 312220.9439 & 1 \\
NHP & \textcolor{nicered}{Context-Aware} & \cellcolor{eYouTube}YouTube & \cellcolor{eNews}news & 1968.7602 & 3.3786 & 1.2290 \\
NHP & \textcolor{nicered}{Context-Free} & \cellcolor{eYouTube}YouTube & \cellcolor{eNews}news & 497.7945 & 2.8012 & 1.0190 \\
NHP & \textcolor{nicegreen}{Context-Aware Data-Aug} & \cellcolor{eYouTube}YouTube & \cellcolor{eNews}news & 78.1916 & 267162.4407 & 0.9525 \\
NHP & \textcolor{nicegreen}{Context-Free Data-Aug} & \cellcolor{eYouTube}YouTube & \cellcolor{eNews}news & 121.1747 & 292707.2868 & 0.8355 \\
NTP & \textcolor{niceblue}{Hypernyms Baseline} & \cellcolor{eYouTube}YouTube & \cellcolor{eNews}news & 92.3580 & 378706.8522 & 1.2129 \\
NHP & \textcolor{nicered}{Context-Aware} & \cellcolor{eYouTube}YouTube & \cellcolor{eArXiv}arxiv & 349.2148 & 2.7853 & 1.0132 \\
NHP & \textcolor{nicered}{Context-Free} & \cellcolor{eYouTube}YouTube & \cellcolor{eArXiv}arxiv & 349.2148 & 2.7853 & 1.0132 \\
NHP & \textcolor{nicegreen}{Context-Aware Data-Aug} & \cellcolor{eYouTube}YouTube & \cellcolor{eArXiv}arxiv & 118.6937 & 420902.7825 & 1.5007 \\
NHP & \textcolor{nicegreen}{Context-Free Data-Aug} & \cellcolor{eYouTube}YouTube & \cellcolor{eArXiv}arxiv & 105.6572 & 270313.4715 & 0.7716 \\
NTP & \textcolor{niceblue}{Hypernyms Baseline} & \cellcolor{eYouTube}YouTube & \cellcolor{eArXiv}arxiv & 103.7129 & 208530.3614 & 0.6679 \\
NHP & \textcolor{nicered}{Context-Aware} & \cellcolor{eYouTube}YouTube & \cellcolor{eCombined}combined & 289.4326 & 2.7863 & 1.0136 \\
NHP & \textcolor{nicered}{Context-Free} & \cellcolor{eYouTube}YouTube & \cellcolor{eCombined}combined & 266.6112 & 2.7370 & 0.9956 \\
NHP & \textcolor{nicegreen}{Context-Aware Data-Aug} & \cellcolor{eYouTube}YouTube & \cellcolor{eCombined}combined & \textbf{66.5035} & 298317.9281 & 1.0636 \\
NHP & \textcolor{nicegreen}{Context-Free Data-Aug} & \cellcolor{eYouTube}YouTube & \cellcolor{eCombined}combined & 76.9409 & 295248.1647 & 0.8428 \\
NTP & \textcolor{niceblue}{Hypernyms Baseline} & \cellcolor{eYouTube}YouTube & \cellcolor{eCombined}combined & 82.6283 & 746022.1188 & 2.3894 \\
\midrule

NHP & \textcolor{nicered}{Context-Aware} & \cellcolor{eCombined}Combined & \cellcolor{eYouTube}youtube & 422.6810 & 2.8459 & 1.0291 \\
NHP & \textcolor{nicered}{Context-Free} & \cellcolor{eCombined}Combined & \cellcolor{eYouTube}youtube & 417.5940 & 2.8459 & 1.0568 \\
NHP & \textcolor{nicegreen}{Context-Aware Data-Aug} & \cellcolor{eCombined}Combined & \cellcolor{eYouTube}youtube & 89.0367 & 200920.7243 & 0.9743 \\
NHP & \textcolor{nicegreen}{Context-Free Data-Aug} & \cellcolor{eCombined}Combined & \cellcolor{eYouTube}youtube & 102.9593 & 254965.1385 & 1.0226 \\
NTP & \textcolor{niceblue}{Hypernyms Baseline} & \cellcolor{eCombined}Combined & \cellcolor{eYouTube}youtube & 115.3129 & 218921.3433 & 0.4557 \\
NHP & \textcolor{nicered}{Context-Aware} & \cellcolor{eCombined}Combined & \cellcolor{eNews}news & 1213.9497 & 3.3974 & 1.2288 \\
NHP & \textcolor{nicered}{Context-Free} & \cellcolor{eCombined}Combined & \cellcolor{eNews}news & 451.0357 & 2.7685 & 1.0282 \\
NHP & \textcolor{nicegreen}{Context-Aware Data-Aug} & \cellcolor{eCombined}Combined & \cellcolor{eNews}news & 66.9841 & 189212.1734 & 0.9176 \\
NHP & \textcolor{nicegreen}{Context-Free Data-Aug} & \cellcolor{eCombined}Combined & \cellcolor{eNews}news & 88.6590 & 192357.5958 & 0.7714 \\
NTP & \textcolor{niceblue}{Hypernyms Baseline} & \cellcolor{eCombined}Combined & \cellcolor{eNews}news & 74.4351 & 271243.1141 & 0.5647 \\
NHP & \textcolor{nicered}{Context-Aware} & \cellcolor{eCombined}Combined & \cellcolor{eArXiv}arxiv & 331.7483 & 2.7873 & 1.0080 \\
NHP & \textcolor{nicered}{Context-Free} & \cellcolor{eCombined}Combined & \cellcolor{eArXiv}arxiv & 331.7483 & 2.7873 & 1.0353 \\
NHP & \textcolor{nicegreen}{Context-Aware Data-Aug} & \cellcolor{eCombined}Combined & \cellcolor{eArXiv}arxiv & 94.0573 & 282551.9262 & 1.3702 \\
NHP & \textcolor{nicegreen}{Context-Free Data-Aug} & \cellcolor{eCombined}Combined & \cellcolor{eArXiv}arxiv & 83.5826 & 191881.4087 & 0.7696 \\
NTP & \textcolor{niceblue}{Hypernyms Baseline} & \cellcolor{eCombined}Combined & \cellcolor{eArXiv}arxiv & 87.6069 & 147708.6478 & 0.3074 \\
NHP & \textcolor{nicered}{Context-Aware} & \cellcolor{eCombined}Combined & \cellcolor{eCombined}combined & 243.7188 & 2.7656 & 1 \\
NHP & \textcolor{nicered}{Context-Free} & \cellcolor{eCombined}Combined & \cellcolor{eCombined}combined & 227.7254 & 2.6928 & 1 \\
NHP & \textcolor{nicegreen}{Context-Aware Data-Aug} & \cellcolor{eCombined}Combined & \cellcolor{eCombined}combined & \textbf{55.5191} & 206195.8661 & 1 \\
NHP & \textcolor{nicegreen}{Context-Free Data-Aug} & \cellcolor{eCombined}Combined & \cellcolor{eCombined}combined & 63.8804 & 249322.4276 & 1 \\
NTP & \textcolor{niceblue}{Hypernyms Baseline} & \cellcolor{eCombined}Combined & \cellcolor{eCombined}combined & 63.2772 & 480380.8561 & 1 \\
\midrule

NHP & \textcolor{nicered}{Context-Aware} & \cellcolor{eArXiv}ArXiv & \cellcolor{eYouTube}youtube & 198.3470 & 3.2236 & 1.0528 \\
NHP & \textcolor{nicered}{Context-Free} & \cellcolor{eArXiv}ArXiv & \cellcolor{eYouTube}youtube & 198.3470 & 3.2197 & 1.0515 \\
NHP & \textcolor{nicegreen}{Context-Aware Data-Aug} & \cellcolor{eArXiv}ArXiv & \cellcolor{eYouTube}youtube & 54.3881 & 142005.4560 & 0.7963 \\
NHP & \textcolor{nicegreen}{Context-Free Data-Aug} & \cellcolor{eArXiv}ArXiv & \cellcolor{eYouTube}youtube & 69.5159 & 176528.8680 & 1.3255 \\
NTP & \textcolor{niceblue}{Hypernyms Baseline} & \cellcolor{eArXiv}ArXiv & \cellcolor{eYouTube}youtube & 83.9096 & 150683.8633 & 1.4585 \\
NHP & \textcolor{nicered}{Context-Aware} & \cellcolor{eArXiv}ArXiv & \cellcolor{eNews}news & 1021.3224 & 4.3435 & 1.4185 \\
NHP & \textcolor{nicered}{Context-Free} & \cellcolor{eArXiv}ArXiv & \cellcolor{eNews}news & 268.4659 & 3.2313 & 1.0553 \\
NHP & \textcolor{nicegreen}{Context-Aware Data-Aug} & \cellcolor{eArXiv}ArXiv & \cellcolor{eNews}news & 46.8513 & 140642.6005 & 0.7886 \\
NHP & \textcolor{nicegreen}{Context-Free Data-Aug} & \cellcolor{eArXiv}ArXiv & \cellcolor{eNews}news & 69.9829 & 137154.0853 & 1.0298 \\
NTP & \textcolor{niceblue}{Hypernyms Baseline} & \cellcolor{eArXiv}ArXiv & \cellcolor{eNews}news & 54.2679 & 194982.2997 & 1.8866 \\
NHP & \textcolor{nicered}{Context-Aware} & \cellcolor{eArXiv}ArXiv & \cellcolor{eArXiv}arxiv & 209.9622 & 3.0619 & 1 \\
NHP & \textcolor{nicered}{Context-Free} & \cellcolor{eArXiv}ArXiv & \cellcolor{eArXiv}arxiv & 209.9622 & 3.0619 & 1 \\
NHP & \textcolor{nicegreen}{Context-Aware Data-Aug} & \cellcolor{eArXiv}ArXiv & \cellcolor{eArXiv}arxiv & 60.9543 & 178357.1354 & 1 \\
NHP & \textcolor{nicegreen}{Context-Free Data-Aug} & \cellcolor{eArXiv}ArXiv & \cellcolor{eArXiv}arxiv & 63.5198 & 133186.0594 & 1 \\
NTP & \textcolor{niceblue}{Hypernyms Baseline} & \cellcolor{eArXiv}ArXiv & \cellcolor{eArXiv}arxiv & 58.0968 & 103344.6714 & 1 \\
NHP & \textcolor{nicered}{Context-Aware} & \cellcolor{eArXiv}ArXiv & \cellcolor{eCombined}combined & 183.4330 & 3.1595 & 1.0319 \\
NHP & \textcolor{nicered}{Context-Free} & \cellcolor{eArXiv}ArXiv & \cellcolor{eCombined}combined & 188.9383 & 3.0557 & 0.9971 \\
NHP & \textcolor{nicegreen}{Context-Aware Data-Aug} & \cellcolor{eArXiv}ArXiv & \cellcolor{eCombined}combined & \textbf{49.1487} & 145251.7602 & 0.8145 \\
NHP & \textcolor{nicegreen}{Context-Free Data-Aug} & \cellcolor{eArXiv}ArXiv & \cellcolor{eCombined}combined & 56.2360 & 206822.9312 & 1.5528 \\
NTP & \textcolor{niceblue}{Hypernyms Baseline} & \cellcolor{eArXiv}ArXiv & \cellcolor{eCombined}combined & 56.6354 & 336169.0554 & 3.2534 \\
\midrule

NHP & \textcolor{nicered}{Context-Aware} & \cellcolor{eNews}News & \cellcolor{eYouTube}youtube & 249.7320 & 2.7484 & 0.9593 \\
NHP & \textcolor{nicered}{Context-Free} & \cellcolor{eNews}News & \cellcolor{eYouTube}youtube & 249.7320 & 2.7484 & 1.1180 \\
NHP & \textcolor{nicegreen}{Context-Aware Data-Aug} & \cellcolor{eNews}News & \cellcolor{eYouTube}youtube & 65.9469 & 194415.6066 & 1.1357 \\
NHP & \textcolor{nicegreen}{Context-Free Data-Aug} & \cellcolor{eNews}News & \cellcolor{eYouTube}youtube & 73.8743 & 251033.5765 & 1.4817 \\
NTP & \textcolor{niceblue}{Hypernyms Baseline} & \cellcolor{eNews}News & \cellcolor{eYouTube}youtube & 79.8546 & 211511.4507 & 0.8334 \\
NHP & \textcolor{nicered}{Context-Aware} & \cellcolor{eNews}News & \cellcolor{eNews}news & 399.0138 & 2.8651 & 1 \\
NHP & \textcolor{nicered}{Context-Free} & \cellcolor{eNews}News & \cellcolor{eNews}news & 189.8203 & 2.4584 & 1 \\
NHP & \textcolor{nicegreen}{Context-Aware Data-Aug} & \cellcolor{eNews}News & \cellcolor{eNews}news & 38.9584 & 171159.6621 & 1 \\
NHP & \textcolor{nicegreen}{Context-Free Data-Aug} & \cellcolor{eNews}News & \cellcolor{eNews}news & 42.1878 & 169404.3654 & 1 \\
NTP & \textcolor{niceblue}{Hypernyms Baseline} & \cellcolor{eNews}News & \cellcolor{eNews}news & 36.9524 & 253761.9440 & 1 \\
NHP & \textcolor{nicered}{Context-Aware} & \cellcolor{eNews}News & \cellcolor{eArXiv}arxiv & 217.1768 & 2.6731 & 0.9328 \\
NHP & \textcolor{nicered}{Context-Free} & \cellcolor{eNews}News & \cellcolor{eArXiv}arxiv & 217.1768 & 2.6731 & 1.0872 \\
NHP & \textcolor{nicegreen}{Context-Aware Data-Aug} & \cellcolor{eNews}News & \cellcolor{eArXiv}arxiv & 65.9476 & 279835.1470 & 1.6351 \\
NHP & \textcolor{nicegreen}{Context-Free Data-Aug} & \cellcolor{eNews}News & \cellcolor{eArXiv}arxiv & 60.1009 & 185352.6254 & 1.0941 \\
NTP & \textcolor{niceblue}{Hypernyms Baseline} & \cellcolor{eNews}News & \cellcolor{eArXiv}arxiv & 63.6622 & 141820.9889 & 0.5587 \\
NHP & \textcolor{nicered}{Context-Aware} & \cellcolor{eNews}News & \cellcolor{eCombined}combined & 155.4530 & 2.5233 & 0.8806 \\
NHP & \textcolor{nicered}{Context-Free} & \cellcolor{eNews}News & \cellcolor{eCombined}combined & 148.9760 & 2.4833 & 1.0101 \\
NHP & \textcolor{nicegreen}{Context-Aware Data-Aug} & \cellcolor{eNews}News & \cellcolor{eCombined}combined & \textbf{34.3552} & 190031.9093 & 1.1100 \\
NHP & \textcolor{nicegreen}{Context-Free Data-Aug} & \cellcolor{eNews}News & \cellcolor{eCombined}combined & 39.1561 & 240334.4371 & 1.4184 \\
NTP & \textcolor{niceblue}{Hypernyms Baseline} & \cellcolor{eNews}News & \cellcolor{eCombined}combined & 36.1536 & 417731.6536 & 1.6462 \\
\midrule

NSP & \textcolor{nicered}{Context-Aware} & \cellcolor{eYouTube}YouTube & \cellcolor{eYouTube}youtube & 1281.9174 & 3.3604 & 1 \\
NSP & \textcolor{nicered}{Context-Free} & \cellcolor{eYouTube}YouTube & \cellcolor{eYouTube}youtube & 1281.9174 & 3.3604 & 1 \\
NSP & \textcolor{nicegreen}{Context-Aware Data-Aug} & \cellcolor{eYouTube}YouTube & \cellcolor{eYouTube}youtube & 257.4089 & 260072.7351 & 1 \\
NSP & \textcolor{nicegreen}{Context-Free Data-Aug} & \cellcolor{eYouTube}YouTube & \cellcolor{eYouTube}youtube & 258.3046 & 278252.2883 & 1 \\
NTP & \textcolor{niceblue}{Synonyms Baseline} & \cellcolor{eYouTube}YouTube & \cellcolor{eYouTube}youtube & 184.3536 & 663891.0267 & 1 \\
NSP & \textcolor{nicered}{Context-Aware} & \cellcolor{eYouTube}YouTube & \cellcolor{eNews}news & 5021.3823 & 3.8017 & 1.1313 \\
NSP & \textcolor{nicered}{Context-Free} & \cellcolor{eYouTube}YouTube & \cellcolor{eNews}news & 5021.3823 & 3.8017 & 1.1313 \\
NSP & \textcolor{nicegreen}{Context-Aware Data-Aug} & \cellcolor{eYouTube}YouTube & \cellcolor{eNews}news & 494.5056 & 321449.3919 & 1.2359 \\
NSP & \textcolor{nicegreen}{Context-Free Data-Aug} & \cellcolor{eYouTube}YouTube & \cellcolor{eNews}news & 553.3879 & 751897.6226 & 2.7022 \\
NTP & \textcolor{niceblue}{Synonyms Baseline} & \cellcolor{eYouTube}YouTube & \cellcolor{eNews}news & 244.5672 & 88239614.03 & 1.3266 \\
NSP & \textcolor{nicered}{Context-Aware} & \cellcolor{eYouTube}YouTube & \cellcolor{eArXiv}arxiv & 3433.2579 & 3.9910 & 1.1876 \\
NSP & \textcolor{nicered}{Context-Free} & \cellcolor{eYouTube}YouTube & \cellcolor{eArXiv}arxiv & 2731.1732 & 4.0168 & 1.1953 \\
NSP & \textcolor{nicegreen}{Context-Aware Data-Aug} & \cellcolor{eYouTube}YouTube & \cellcolor{eArXiv}arxiv & 448.1877 & 948690.9478 & 3.6477 \\
NSP & \textcolor{nicegreen}{Context-Free Data-Aug} & \cellcolor{eYouTube}YouTube & \cellcolor{eArXiv}arxiv & 523.8087 & 504478.2195 & 1.8130 \\
NTP & \textcolor{niceblue}{Synonyms Baseline} & \cellcolor{eYouTube}YouTube & \cellcolor{eArXiv}arxiv & 841.5366 & 508670.5981 & 4.5647 \\
NSP & \textcolor{nicered}{Context-Aware} & \cellcolor{eYouTube}YouTube & \cellcolor{eCombined}combined & 1364.8438 & 315.4954 & 93.8862 \\
NSP & \textcolor{nicered}{Context-Free} & \cellcolor{eYouTube}YouTube & \cellcolor{eCombined}combined & 9910.6903 & 315.4954 & 93.8862 \\
NSP & \textcolor{nicegreen}{Context-Aware Data-Aug} & \cellcolor{eYouTube}YouTube & \cellcolor{eCombined}combined & 194.8153 & 367184.1088 & 1.4118 \\
NSP & \textcolor{nicegreen}{Context-Free Data-Aug} & \cellcolor{eYouTube}YouTube & \cellcolor{eCombined}combined & 192.9121 & 1415625.616 & 5.0876 \\
NTP & \textcolor{niceblue}{Synonyms Baseline} & \cellcolor{eYouTube}YouTube & \cellcolor{eCombined}combined & 162.7477 & 2243779.964 & 0.8828 \\
\midrule

NSP & \textcolor{nicered}{Context-Aware} & \cellcolor{eCombined}Combined & \cellcolor{eYouTube}youtube & 1675.9424 & 4.1230 & 0.0153 \\
NSP & \textcolor{nicered}{Context-Free} & \cellcolor{eCombined}Combined & \cellcolor{eYouTube}youtube & 1675.9424 & 4.1230 & 0.0153 \\
NSP & \textcolor{nicegreen}{Context-Aware Data-Aug} & \cellcolor{eCombined}Combined & \cellcolor{eYouTube}youtube & 297.2277 & 198108.5509 & 0.7760 \\
NSP & \textcolor{nicegreen}{Context-Free Data-Aug} & \cellcolor{eCombined}Combined & \cellcolor{eYouTube}youtube & 307.2453 & 222821.4406 & 0.2394 \\
NSP & \textcolor{niceblue}{Synonyms Baseline} & \cellcolor{eCombined}Combined & \cellcolor{eYouTube}youtube & 209.5685 & 516784.9742 & 1.7082 \\
NSP & \textcolor{nicered}{Context-Aware} & \cellcolor{eCombined}Combined & \cellcolor{eNews}news & 3704.0467 & 4.1497 & 0.0154 \\
NSP & \textcolor{nicered}{Context-Free} & \cellcolor{eCombined}Combined & \cellcolor{eNews}news & 3704.0467 & 4.1497 & 0.0154 \\
NSP & \textcolor{nicegreen}{Context-Aware Data-Aug} & \cellcolor{eCombined}Combined & \cellcolor{eNews}news & 289.5167 & 228742.4686 & 0.8961 \\
NSP & \textcolor{nicegreen}{Context-Free Data-Aug} & \cellcolor{eCombined}Combined & \cellcolor{eNews}news & 317.2183 & 499541.8261 & 0.5368 \\
NTP & \textcolor{niceblue}{Synonyms Baseline} & \cellcolor{eCombined}Combined & \cellcolor{eNews}news & 159.1297 & 62285244.35 & 3.7888 \\
NSP & \textcolor{nicered}{Context-Aware} & \cellcolor{eCombined}Combined & \cellcolor{eArXiv}arxiv & 1771.6198 & 4.1760 & 0.0155 \\
NSP & \textcolor{nicered}{Context-Free} & \cellcolor{eCombined}Combined & \cellcolor{eArXiv}arxiv & 1785.0295 & 4.2526 & 0.0158 \\
NSP & \textcolor{nicegreen}{Context-Aware Data-Aug} & \cellcolor{eCombined}Combined & \cellcolor{eArXiv}arxiv & 349.5710 & 639379.0527 & 2.5047 \\
NSP & \textcolor{nicegreen}{Context-Free Data-Aug} & \cellcolor{eCombined}Combined & \cellcolor{eArXiv}arxiv & 435.6399 & 380467.2669 & 0.4088 \\
NTP & \textcolor{niceblue}{Synonyms Baseline} & \cellcolor{eCombined}Combined & \cellcolor{eArXiv}arxiv & 471.3260 & 358696.7699 & 3.8419 \\
NSP & \textcolor{nicered}{Context-Aware} & \cellcolor{eCombined}Combined & \cellcolor{eCombined}combined & 2717.3075 & 269.3490 & 1 \\
NSP & \textcolor{nicered}{Context-Free} & \cellcolor{eCombined}Combined & \cellcolor{eCombined}combined & 2717.3075 & 269.3490 & 1 \\
NSP & \textcolor{nicegreen}{Context-Aware Data-Aug} & \cellcolor{eCombined}Combined & \cellcolor{eCombined}combined & 141.0141 & 255268.2881 & 1 \\
NSP & \textcolor{nicegreen}{Context-Free Data-Aug} & \cellcolor{eCombined}Combined & \cellcolor{eCombined}combined & 130.5401 & 930607.8148 & 1 \\
NTP & \textcolor{niceblue}{Synonyms Baseline} & \cellcolor{eCombined}Combined & \cellcolor{eCombined}combined & 122.6778 & 1673866.06 & 1 \\
\midrule

NSP & \textcolor{nicered}{Context-Aware} & \cellcolor{eArXiv}ArXiv & \cellcolor{eYouTube}youtube & 1776.9550 & 5.5599 & 1.2030 \\
NSP & \textcolor{nicered}{Context-Free} & \cellcolor{eArXiv}ArXiv & \cellcolor{eYouTube}youtube & 1776.9550 & 5.5599 & 1.1501 \\
NSP & \textcolor{nicegreen}{Context-Aware Data-Aug} & \cellcolor{eArXiv}ArXiv & \cellcolor{eYouTube}youtube & 248.6060 & 143742.8271 & 0.3201 \\
NSP & \textcolor{nicegreen}{Context-Free Data-Aug} & \cellcolor{eArXiv}ArXiv & \cellcolor{eYouTube}youtube & 308.6261 & 172417.5462 & 0.5997 \\
NTP & \textcolor{niceblue}{Synonyms Baseline} & \cellcolor{eArXiv}ArXiv & \cellcolor{eYouTube}youtube & 170.6700 & 371409.9797 & 0.3079 \\
NSP & \textcolor{nicered}{Context-Aware} & \cellcolor{eArXiv}ArXiv & \cellcolor{eNews}news & 4027.9929 & 5.5573 & 1.2025 \\
NSP & \textcolor{nicered}{Context-Free} & \cellcolor{eArXiv}ArXiv & \cellcolor{eNews}news & 4027.9929 & 5.5573 & 1.1495 \\
NSP & \textcolor{nicegreen}{Context-Aware Data-Aug} & \cellcolor{eArXiv}ArXiv & \cellcolor{eNews}news & 267.5049 & 168677.4707 & 0.3756 \\
NSP & \textcolor{nicegreen}{Context-Free Data-Aug} & \cellcolor{eArXiv}ArXiv & \cellcolor{eNews}news & 298.6867 & 344438.8796 & 1.1981 \\
NTP & \textcolor{niceblue}{Synonyms Baseline} & \cellcolor{eArXiv}ArXiv & \cellcolor{eNews}news & 112.5678 & 36702868.79 & 1.9408 \\
NSP & \textcolor{nicered}{Context-Aware} & \cellcolor{eArXiv}ArXiv & \cellcolor{eArXiv}arxiv & 1002.6701 & 4.6216 & 1 \\
NSP & \textcolor{nicered}{Context-Free} & \cellcolor{eArXiv}ArXiv & \cellcolor{eArXiv}arxiv & 1142.1247 & 4.8344 & 1 \\
NSP & \textcolor{nicegreen}{Context-Aware Data-Aug} & \cellcolor{eArXiv}ArXiv & \cellcolor{eArXiv}arxiv & 316.4381 & 449111.7608 & 1 \\
NSP & \textcolor{nicegreen}{Context-Free Data-Aug} & \cellcolor{eArXiv}ArXiv & \cellcolor{eArXiv}arxiv & 372.3505 & 287483.5986 & 1 \\
NTP & \textcolor{niceblue}{Synonyms Baseline} & \cellcolor{eArXiv}ArXiv & \cellcolor{eArXiv}arxiv & 554.2414 & 254618.3941 & 1 \\
NSP & \textcolor{nicered}{Context-Aware} & \cellcolor{eArXiv}ArXiv & \cellcolor{eCombined}combined & 683.2708 & 227.0362 & 49.1250 \\
NSP & \textcolor{nicered}{Context-Free} & \cellcolor{eArXiv}ArXiv & \cellcolor{eCombined}combined & 683.2708 & 227.0362 & 46.9626 \\
NSP & \textcolor{nicegreen}{Context-Aware Data-Aug} & \cellcolor{eArXiv}ArXiv & \cellcolor{eCombined}combined & 147.8072 & 185486.1319 & 0.4130 \\
NSP & \textcolor{nicegreen}{Context-Free Data-Aug} & \cellcolor{eArXiv}ArXiv & \cellcolor{eCombined}combined & 104.4113 & 579093.5982 & 2.0144 \\
NTP & \textcolor{niceblue}{Synonyms Baseline} & \cellcolor{eArXiv}ArXiv & \cellcolor{eCombined}combined & 121.6163 & 1201784.535 & 0.2194 \\
\midrule

NSP & \textcolor{nicered}{Context-Aware} & \cellcolor{eNews}News & \cellcolor{eYouTube}youtube & 1543.8018 & 4.0254 & 1.1013 \\
NSP & \textcolor{nicered}{Context-Free} & \cellcolor{eNews}News & \cellcolor{eYouTube}youtube & 1543.8018 & 4.0254 & 1.1013 \\
NSP & \textcolor{nicegreen}{Context-Aware Data-Aug} & \cellcolor{eNews}News & \cellcolor{eYouTube}youtube & 255.1020 & 200136.8360 & 0.9257 \\
NSP & \textcolor{nicegreen}{Context-Free Data-Aug} & \cellcolor{eNews}News & \cellcolor{eYouTube}youtube & 246.7716 & 222645.3484 & 0.4730 \\
NTP & \textcolor{niceblue}{Synonyms Baseline} & \cellcolor{eNews}News & \cellcolor{eYouTube}youtube & 163.8084 & 525066.3462 & 0.6539 \\
NSP & \textcolor{nicered}{Context-Aware} & \cellcolor{eNews}News & \cellcolor{eNews}news & 1680.7794 & 3.6551 & 1 \\
NSP & \textcolor{nicered}{Context-Free} & \cellcolor{eNews}News & \cellcolor{eNews}news & 1680.7794 & 3.6551 & 1 \\
NSP & \textcolor{nicegreen}{Context-Aware Data-Aug} & \cellcolor{eNews}News & \cellcolor{eNews}news & 171.6298 & 216192.4155 & 1 \\
NSP & \textcolor{nicegreen}{Context-Free Data-Aug} & \cellcolor{eNews}News & \cellcolor{eNews}news & 205.1227 & 470666.8797 & 1 \\
NTP & \textcolor{niceblue}{Synonyms Baseline} & \cellcolor{eNews}News & \cellcolor{eNews}news & 250.4959 & 67641981.74 & 1 \\
NSP & \textcolor{nicered}{Context-Aware} & \cellcolor{eNews}News & \cellcolor{eArXiv}arxiv & 1870.5691 & 4.2061 & 1.1507 \\
NSP & \textcolor{nicered}{Context-Free} & \cellcolor{eNews}News & \cellcolor{eArXiv}arxiv & 1602.5725 & 4.2067 & 1.1509 \\
NSP & \textcolor{nicegreen}{Context-Aware Data-Aug} & \cellcolor{eNews}News & \cellcolor{eArXiv}arxiv & 213.5528 & 606818.9040 & 2.8068 \\
NSP & \textcolor{nicegreen}{Context-Free Data-Aug} & \cellcolor{eNews}News & \cellcolor{eArXiv}arxiv & 299.1280 & 366432.3553 & 0.7785 \\
NTP & \textcolor{niceblue}{Synonyms Baseline} & \cellcolor{eNews}News & \cellcolor{eArXiv}arxiv & 371.2635 & 343182.1046 & 1.4821 \\
NSP & \textcolor{nicered}{Context-Aware} & \cellcolor{eNews}News & \cellcolor{eCombined}combined & 1905.4865 & 275.4430 & 75.3585 \\
NSP & \textcolor{nicered}{Context-Free} & \cellcolor{eNews}News & \cellcolor{eCombined}combined & 1905.4865 & 275.4430 & 0.0013 \\
NSP & \textcolor{nicegreen}{Context-Aware Data-Aug} & \cellcolor{eNews}News & \cellcolor{eCombined}combined & 102.0031 & 236488.3831 & 0.5025 \\
NSP & \textcolor{nicegreen}{Context-Free Data-Aug} & \cellcolor{eNews}News & \cellcolor{eCombined}combined & 79.9511 & 957988.1020 & 0.0142 \\
NTP & \textcolor{niceblue}{Synonyms Baseline} & \cellcolor{eNews}News & \cellcolor{eCombined}combined & 85.6417 & 1600376.4410 & 0.0458 \\
\label{tab:perp_full}
\end{longtable}
}

\begin{table}[h!]
\centering
\label{tab:pretrain_transfer_no_ratios}
\begin{tabular}{ll|c}
\hline
\hline
\textbf{Train\ \ \ \ \ } & \textbf{Eval} & \textbf{Best Model} \\
\hline
\hline
\multirow{5}{*}{\parbox{1pt}{\textbf{YouTube}}}
    & YouTube & {\color{niceblue}NTP Synonym Baseline}\\
    \cdashline{2-3}
    & News & \makecell{\color{nicegreen}NHP Context-\\\color{nicegreen}Aware Data Aug.}\\
    \cdashline{2-3}
    & ArXiv & \makecell{\color{nicegreen}NHP Context-\\\color{nicegreen}Aware Data Aug.}\\
    \cdashline{2-3}
    & Combined & \makecell{\color{nicegreen}NHP Context-\\\color{nicegreen}Aware Data Aug.}\\
\hline
\multirow{5}{*}{\textbf{News}} 
    & News & {\color{niceblue}NTP Synonym Baseline}\\
    \cdashline{2-3}
    & YouTube &  \makecell{\color{nicegreen}NHP Context-\\\color{nicegreen}Aware Data Aug.}\\
    \cdashline{2-3}
    & ArXiv & \makecell{\color{nicegreen}NHP Context-\\\color{nicegreen}Aware Data Aug.}\\
    \cdashline{2-3}
    & Combined & \makecell{\color{nicegreen}NHP Context-\\\color{nicegreen}Aware Data Aug.}\\
\hline
\multirow{5}{*}{\textbf{ArXiv}} 
    & ArXiv & {\color{niceblue}NTP Synonym Baseline}\\
    \cdashline{2-3}
    & YouTube &  {\color{niceblue}NTP Synonym Baseline}\\
    \cdashline{2-3}
    & News & \makecell{\color{nicegreen}NHP Context-\\\color{nicegreen}Free Data Aug.}\\
    \cdashline{2-3}
    & Combined &  \makecell{\color{nicegreen}NHP Context-\\\color{nicegreen}Free Data Aug.}\\
\hline
\multirow{5}{*}{Combined} 
    & Combined & \makecell{\color{nicegreen}NHP Context-\\\color{nicegreen}Aware Data Aug.}\\
    \cdashline{2-3}
    & YouTube &  \makecell{\color{nicegreen}NHP Context-\\\color{nicegreen}Aware Data Aug.}\\
    \cdashline{2-3}
    & News & \makecell{\color{nicegreen}NHP Context-\\\color{nicegreen}Aware Data Aug.}\\
    \cdashline{2-3}
    & ArXiv &  \makecell{\color{nicegreen}NHP Context-\\\color{nicegreen}Aware Data Aug.}
\end{tabular}
\caption{Best model using NTP perplexity scores (no ratios, unlike Table 2).}
\end{table}

\section{F1, Precision, Recall, and AUC-PR}
In addition to the main-text F1 scores, we include Precision, Recall, and area under the Precision–Recall curve (AUC-PR) to provide a more complete view of model behavior, particularly under class imbalance. All tables follow the same experimental setup and ordering as in the main results, differing only in the evaluation metric reported. Together, these metrics allow for a finer-grained analysis of trade-offs between false positives and false negatives and help contextualize performance differences across objectives, data augmentation strategies, and training domains.

\begin{table*}[h!]
\centering
\small
\caption{\textbf{Downstream fine-tuned F1 scores across seven benchmarks.}
Precision is reported for \textsc{Empathetic Dialogues} (EMO), \textsc{GLUE}, \textsc{Hate Speech} (HATE), \textsc{SNLI}, \textsc{SpamAssassin} (SPAM), \textsc{Fake News} (FAKE), \textsc{Logical Fallacy} (LOG).}
\label{tab:downstream-all-f1}
\resizebox{\textwidth}{!}{
\begin{tabular}{ll|rrrrrrr}
\hline\hline
\textbf{Variant} & \textbf{Domain} & \textbf{EMO} & \textbf{GLUE} & \textbf{HATE} & \textbf{SNLI} & \textbf{SPAM} & \textbf{FAKE} & \textbf{LOG} \\
\hline\hline

\multirow{4}{*}{\makecell{\color{nicered}NSP Loss\\\color{nicered}Context-Aware}}
& ArXiv    & 0.7961 & 0.7547 & 0.5489 & 0.8134 & 0.9943 & 0.2399 & 0.4398 \\
\cdashline{2-9}
& News     & 0.7539 & 0.7864 & 0.6523 & 0.1758 & 0.9641 & 0.4780 & 0.4024 \\
\cdashline{2-9}
& YouTube  & 0.7957 & 0.8360 & 0.6463 & 0.5094 & 0.9258 & 0.3795 & 0.4892 \\
\cdashline{2-9}
& Combined & 0.8067 & 0.1679 & 0.7128 & 0.1734 & 0.9782 & 0.0000 & 0.4686 \\

\hline
\multirow{4}{*}{\makecell{\color{nicered}NSP Loss\\\color{nicered}Context-Free}}
& ArXiv    & 0.7924 & 0.8383 & 0.5490 & 0.4229 & 0.9570 & 0.1982 & 0.3857 \\
\cdashline{2-9}
& News     & 0.8187 & 0.8273 & 0.5999 & 0.8044 & 0.9641 & 0.0000 & 0.4315 \\
\cdashline{2-9}
& YouTube  & 0.7977 & 0.7819 & 0.5822 & 0.1665 & 0.9745 & 0.0000 & 0.4892 \\
\cdashline{2-9}
& Combined & 0.6189 & 0.5821 & 0.2911 & 0.1665 & 0.9785 & 0.5568 & 0.0159 \\

\hline
\multirow{4}{*}{\makecell{\color{nicered}NHP Loss\\\color{nicered}Context-Aware}}
& ArXiv    & 0.7664 & 0.8451 & 0.3252 & 0.6883 & 0.9932 & 0.6037 & 0.5143 \\
\cdashline{2-9}
& News     & 0.7211 & 0.7541 & 0.4045 & 0.8210 & 0.9710 & 0.5579 & 0.4676 \\
\cdashline{2-9}
& YouTube  & 0.7381 & 0.1637 & 0.3021 & 0.7885 & 0.9659 & 0.3750 & 0.4253 \\
\cdashline{2-9}
& Combined & 0.7805 & 0.8094 & 0.6709 & 0.7490 & 0.9784 & 0.3121 & 0.4620 \\

\hline
\multirow{4}{*}{\makecell{\color{nicered}NHP Loss\\\color{nicered}Context-Free}}
& ArXiv    & 0.8113 & 0.8451 & 0.3252 & 0.6883 & 0.9932 & 0.6037 & 0.5143 \\
\cdashline{2-9}
& News     & 0.7708 & 0.8144 & 0.6009 & 0.1734 & 0.9819 & 0.2439 & 0.4039 \\
\cdashline{2-9}
& YouTube  & 0.7029 & 0.1637 & 0.5493 & 0.7885 & 0.9659 & 0.3750 & 0.4253 \\
\cdashline{2-9}
& Combined & 0.7041 & 0.8160 & 0.6013 & 0.8350 & 0.9338 & 0.3034 & 0.4783 \\

\hline
\multirow{4}{*}{\makecell{\color{nicegreen}NSP Context-\\\color{nicegreen}Aware Data Aug.}}
& ArXiv    & 0.7982 & 0.7583 & 0.7055 & 0.8373 & 0.9821 & 0.2654 & 0.4502 \\
\cdashline{2-9}
& News     & 0.8103 & 0.6097 & 0.4837 & 0.8200 & \textbf{0.9954} & 0.3409 & 0.0420 \\
\cdashline{2-9}
& YouTube  & 0.7206 & 0.7762 & 0.6461 & 0.8278 & 0.9657 & 0.3464 & 0.4585 \\
\cdashline{2-9}
& Combined & 0.7520 & 0.8353 & 0.5830 & 0.8241 & 0.9455 & 0.5380 & 0.4625 \\

\hline
\multirow{4}{*}{\makecell{\color{nicegreen}NSP Context-\\\color{nicegreen}Free Data Aug.}}
& ArXiv    & 0.7982 & 0.5536 & 0.7055 & 0.7955 & 0.9744 & 0.2254 & 0.2832 \\
\cdashline{2-9}
& News     & 0.8198 & 0.8014 & 0.4776 & 0.8507 & 0.9876 & 0.5934 & \textbf{0.5184} \\
\cdashline{2-9}
& YouTube  & 0.7945 & 0.8008 & 0.5422 & 0.8307 & 0.9843 & 0.2860 & 0.4517 \\
\cdashline{2-9}
& Combined & 0.7399 & \textbf{0.8491} & 0.6084 & 0.8622 & 0.9886 & 0.3892 & 0.4941 \\

\hline
\multirow{4}{*}{\makecell{\color{nicegreen}NHP Context-\\\color{nicegreen}Aware Data Aug.}}
& ArXiv    & 0.7538 & 0.3386 & 0.5823 & 0.8250 & 0.9899 & 0.0343 & 0.4473 \\
\cdashline{2-9}
& News     & 0.7871 & 0.8106 & 0.3444 & 0.6908 & 0.9920 & 0.4704 & 0.4835 \\
\cdashline{2-9}
& YouTube  & 0.7206 & 0.8327 & 0.6008 & 0.8482 & 0.8616 & 0.2585 & 0.4417 \\
\cdashline{2-9}
& Combined & 0.7448 & 0.8153 & 0.5948 & 0.8633 & 0.9510 & 0.4231 & 0.4301 \\

\hline
\multirow{4}{*}{\makecell{\color{nicegreen}NHP Context-\\\color{nicegreen}Free Data Aug.}}
& ArXiv    & \textbf{0.8238} & 0.6647 & 0.7396 & 0.7853 & 0.9920 & 0.4483 & 0.4390 \\
\cdashline{2-9}
& News     & 0.7928 & 0.7722 & 0.5985 & 0.8188 & 0.9875 & 0.3644 & 0.4578 \\
\cdashline{2-9}
& YouTube  & 0.7744 & 0.5722 & 0.6396 & 0.1734 & 0.9921 & 0.3821 & 0.4550 \\
\cdashline{2-9}
& Combined & 0.6921 & 0.8289 & \textbf{0.7592} & 0.8498 & 0.9841 & 0.1568 & 0.4125 \\

\hline\hline
\multirow{4}{*}{\makecell{\color{niceblue}NTP Synonym\\\color{niceblue}Baseline Fine-Tuned}}
& ArXiv    & 0.7737 & 0.7354 & 0.6002 & 0.8385 & 0.9370 & 0.3559 & 0.4013 \\
\cdashline{2-9}
& News     & 0.6133 & 0.2340 & 0.4737 & 0.8262 & 0.9426 & 0.3431 & 0.4373 \\
\cdashline{2-9}
& YouTube  & 0.7456 & 0.4487 & 0.5747 & 0.7979 & 0.9921 & 0.1960 & 0.4033 \\
\cdashline{2-9}
& Combined & 0.7849 & 0.7206 & 0.5345 & 0.8530 & 0.9600 & 0.4473 & 0.4922 \\

\hline
\multirow{4}{*}{\makecell{\color{niceblue}NTP Hypernym\\\color{niceblue}Baseline Fine-Tuned}}
& ArXiv    & 0.7904 & 0.6620 & 0.6610 & 0.8397 & 0.9820 & 0.0799 & 0.4242 \\
\cdashline{2-9}
& News     & 0.8193 & 0.8209 & 0.5625 & 0.7684 & 0.9819 & 0.5915 & 0.4397 \\
\cdashline{2-9}
& YouTube  & 0.7804 & 0.8346 & 0.7218 & 0.8137 & 0.9910 & \textbf{0.6347} & 0.4864 \\
\cdashline{2-9}
& Combined & 0.7061 & 0.6084 & 0.6822 & \textbf{0.8658} & 0.9909 & 0.5905 & 0.4282 \\
\hline
\makecell{\color{niceblue}Base Model\\\color{niceblue}Fine-Tuned} & - & 0.7274 & 0.1679 & 0.4078 & 0.7749 & 0.9670 & 0.0695 & 0.4125 \\
\hline
\makecell{\color{niceblue}Base Model} & - & 0.5370 & 0.1679 & 0.3770 & 0.1734 & 0.0000 & 0.0000 & 0.1540 \\
\hline
\end{tabular}}
\end{table*}

\begin{table*}[h!]
\centering
\small
\caption{\textbf{Downstream fine-tuned precision scores across seven benchmarks.}
Precision is reported for \textsc{Empathetic Dialogues} (EMO), \textsc{GLUE}, \textsc{Hate Speech} (HATE), \textsc{SNLI}, \textsc{SpamAssassin} (SPAM), \textsc{Fake News} (FAKE), \textsc{Logical Fallacy} (LOG).}
\label{tab:downstream-all-precision}
\resizebox{\textwidth}{!}{
\begin{tabular}{ll|rrrrrrr}
\hline\hline
\textbf{Variant} & \textbf{Domain} & \textbf{EMO} & \textbf{GLUE} & \textbf{HATE} & \textbf{SNLI} & \textbf{SPAM} & \textbf{FAKE} & \textbf{LOG} \\
\hline\hline

\multirow{4}{*}{\makecell{\color{nicered}NSP Loss\\\color{nicered}Context-Aware}}
& ArXiv    & 0.7974 & 0.7634 & 0.5867 & 0.8247 & 0.9954 & 0.5459 & 0.5250 \\
& News     & 0.7516 & 0.7945 & 0.7285 & 0.3396 & 0.9811 & 0.4757 & 0.5442 \\
& YouTube  & 0.8044 & 0.8356 & 0.6808 & 0.6086 & 0.8653 & 0.5227 & 0.6189 \\
& Combined & 0.8003 & 0.1122 & 0.7098 & 0.1172 & 0.9839 & 0.0000 & 0.5471 \\
\hline

\multirow{4}{*}{\makecell{\color{nicered}NSP Loss\\\color{nicered}Context-Free}}
& ArXiv    & 0.8059 & 0.8380 & 0.7260 & 0.5121 & 0.9506 & 0.5752 & 0.5021 \\
& News     & 0.4740 & 0.7945 & 0.4717 & 0.8170 & 0.9798 & 0.4740 & 0.5943 \\
& YouTube  & 0.8050 & 0.7958 & 0.6791 & 0.1110 & 0.9929 & 0.0000 & 0.6189 \\
& Combined & 0.7454 & 0.6779 & 0.2584 & 0.1110 & 0.9730 & 0.4297 & 0.0089 \\
\hline

\multirow{4}{*}{\makecell{\color{nicered}NHP Loss\\\color{nicered}Context-Aware}}
& ArXiv    & 0.7539 & 0.8479 & 0.6036 & 0.7060 & 0.9887 & 0.4324 & 0.5539 \\
& News     & 0.7959 & 0.8030 & 0.6573 & 0.8228 & 0.9499 & 0.4331 & 0.5965 \\
& YouTube  & 0.7868 & 0.1085 & 0.5183 & 0.8091 & 0.9976 & 0.4263 & 0.5522 \\
& Combined & 0.8020 & 0.8182 & 0.7090 & 0.7500 & 0.9773 & 0.5347 & 0.5255 \\
\hline

\multirow{4}{*}{\makecell{\color{nicered}NHP Loss\\\color{nicered}Context-Free}}
& ArXiv    & 0.8179 & 0.8479 & 0.6036 & 0.7060 & 0.9887 & 0.4324 & 0.5539 \\
& News     & 0.7765 & 0.8166 & 0.5779 & 0.1172 & 0.9732 & 0.5529 & 0.4681 \\
& YouTube  & 0.7776 & 0.5795 & 0.7134 & 0.1172 & 0.9865 & 0.4502 & 0.5786 \\
& Combined & 0.7482 & 0.8162 & 0.6610 & 0.8347 & 0.9204 & 0.6282 & 0.5905 \\
\hline

\multirow{4}{*}{\makecell{\color{nicegreen}NSP Context-\\\color{nicegreen}Aware Data Aug.}}
& ArXiv    & 0.7783 & 0.7778 & 0.6965 & 0.8491 & 0.9648 & 0.5721 & 0.4809 \\
& News     & 0.8117 & 0.6986 & 0.5640 & 0.8290 & 1.0000 & 0.5109 & 0.0339 \\
& YouTube  & 0.7441 & 0.7974 & 0.6667 & 0.8300 & 0.9397 & 0.5249 & 0.5470 \\
& Combined & 0.7878 & 0.8367 & 0.8089 & 0.8247 & 0.9852 & 0.5741 & 0.5899 \\
\hline

\multirow{4}{*}{\makecell{\color{nicegreen}NSP Context-\\\color{nicegreen}Free Data Aug.}}
& ArXiv    & 0.7973 & 0.6169 & 0.6940 & 0.7998 & 0.9522 & 0.5532 & 0.3603 \\
& News     & 0.8302 & 0.8029 & 0.4717 & 0.8530 & 0.9798 & 0.4740 & 0.6136 \\
& YouTube  & 0.8094 & 0.8040 & 0.6819 & 0.8336 & 0.9691 & 0.5294 & 0.5463 \\
& Combined & 0.7943 & \textbf{0.8494} & 0.7191 & 0.8691 & 0.9908 & 0.4500 & 0.5953 \\
\hline

\multirow{4}{*}{\makecell{\color{nicegreen}NHP Context-\\\color{nicegreen}Aware Data Aug.}}
& ArXiv    & 0.7809 & 0.3602 & 0.7620 & 0.8329 & 0.9799 & 0.5417 & 0.5424 \\
& News     & 0.7944 & 0.8104 & 0.6317 & 0.7215 & \textbf{0.9977} & 0.5406 & 0.6360 \\
& YouTube  & 0.7735 & 0.8412 & 0.6362 & 0.8522 & 0.9970 & 0.5837 & 0.5419 \\
& Combined & 0.7971 & 0.8336 & 0.8240 & 0.8635 & 1.0000 & 0.5917 & 0.5654 \\
\hline

\multirow{4}{*}{\makecell{\color{nicegreen}NHP Context-\\\color{nicegreen}Free Data Aug.}}
& ArXiv    & \textbf{0.8308} & 0.7080 & 0.7397 & 0.7972 & \textbf{0.9977} & 0.4992 & 0.5406 \\
& News     & 0.7972 & 0.7794 & 0.5678 & 0.8327 & 0.9820 & 0.4908 & 0.6063 \\
& YouTube  & 0.7901 & 0.5795 & 0.6918 & 0.1172 & 0.9865 & 0.4502 & 0.5786 \\
& Combined & 0.7658 & 0.8358 & 0.7376 & 0.8526 & 0.9841 & 0.4759 & 0.4806 \\
\hline \hline
\multirow{4}{*}{\makecell{\color{niceblue}NTP Synonym\\\color{niceblue}Baseline Fine-Tuned}}
& ArXiv    & 0.7783 & 0.7792 & 0.5686 & 0.8397 & 0.9801 & 0.5780 & \textbf{0.6383} \\
& News     & 0.7618 & 0.2278 & \textbf{0.8780} & 0.8268 & 0.9924 & 0.6140 & 0.5844 \\
& YouTube  & 0.8058 & 0.5423 & 0.5658 & 0.8130 & 0.9887 & 0.5145 & 0.4998 \\
& Combined & 0.7883 & 0.7639 & 0.7125 & 0.8581 & 0.9371 & 0.5361 & 0.5670 \\
\hline

\multirow{4}{*}{\makecell{\color{niceblue}NTP Hypernym\\\color{niceblue}Baseline Fine-Tuned}}
& ArXiv    & 0.7948 & 0.7310 & 0.7597 & 0.8412 & 0.9710 & \textbf{0.7561} & 0.4712 \\
& News     & 0.8227 & 0.8272 & 0.6993 & 0.7868 & 0.9499 & 0.4646 & 0.4970 \\
& YouTube  & 0.7929 & 0.8345 & 0.7486 & 0.8241 & 0.9821 & 0.4756 & 0.5406 \\
& Combined & 0.7519 & 0.6344 & 0.7098 & \textbf{0.8696} & 0.9865 & 0.4968 & 0.4764 \\
\hline
\makecell{\color{niceblue}Base Model\\\color{niceblue}Fine-Tuned} & - & 0.7585 & 0.1122 & 0.5532 & 0.7842 & 0.9360 & 0.6429 & 0.5527 \\
\hline
\makecell{\color{niceblue}Base Model} & - & 0.3100 & 0.1122 & 0.2000 & 0.1172 & 0.0000 & 0.0000 & 0.0860 \\
\hline

\end{tabular}}
\end{table*}

\begin{table*}[h!]
\centering
\small
\caption{\textbf{Downstream fine-tuned recall scores across seven benchmarks.}
Precision is reported for \textsc{Empathetic Dialogues} (EMO), \textsc{GLUE}, \textsc{Hate Speech} (HATE), \textsc{SNLI}, \textsc{SpamAssassin} (SPAM), \textsc{Fake News} (FAKE), \textsc{Logical Fallacy} (LOG).}
\label{tab:downstream-recall}
\resizebox{\textwidth}{!}{
\begin{tabular}{ll|rrrrrrr}
\hline\hline
\textbf{Variant} & \textbf{Domain} & \textbf{EMO} & \textbf{GLUE} & \textbf{HATE} & \textbf{SNLI} & \textbf{SPAM} & \textbf{FAKE} & \textbf{LOG} \\
\hline\hline

\multirow{4}{*}{\makecell{\color{nicered}NSP Loss\\\color{nicered}Context-Aware}}
& ArXiv    & 0.7986 & 0.7542 & 0.5541 & 0.8115 & 0.9932 & 0.1537 & 0.4306 \\
& News     & 0.7565 & 0.7848 & 0.6465 & 0.3345 & 0.9476 & 0.4803 & 0.4087 \\
& YouTube  & 0.7910 & 0.8372 & 0.6526 & 0.5647 & 0.9954 & 0.2980 & 0.4729 \\
& Combined & 0.8166 & 0.3333 & 0.7210 & 0.3333 & 0.9727 & 0.0000 & 0.4582 \\
\hline

\multirow{4}{*}{\makecell{\color{nicered}NSP Loss\\\color{nicered}Context-Free}}
& ArXiv    & 0.7847 & 0.8392 & 0.5196 & 0.4748 & 0.9636 & 0.1197 & 0.3910 \\
& News     & 0.4803 & 0.8284 & 0.4850 & 0.3333 & 0.0000 & 0.7932 & \textbf{0.5034} \\
& YouTube  & 0.7929 & 0.7789 & 0.5526 & 0.3333 & 0.9567 & 0.0000 & 0.4729 \\
& Combined & 0.6049 & 0.6045 & 0.3333 & 0.3333 & 0.9841 & 0.7905 & 0.0769 \\
\hline

\multirow{4}{*}{\makecell{\color{nicered}NHP Loss\\\color{nicered}Context-Aware}}
& ArXiv    & 0.7412 & \textbf{0.8438} & 0.6048 & 0.8227 & \textbf{1.0000} & 0.0177 & 0.4396 \\
& News     & 0.7833 & 0.7643 & 0.3957 & 0.8218 & 0.9932 & 0.7837 & 0.4623 \\
& YouTube  & 0.7118 & 0.3333 & 0.3384 & 0.7856 & 0.9362 & 0.3347 & 0.4327 \\
& Combined & 0.7227 & 0.8070 & 0.5852 & 0.7552 & 0.9066 & 0.2204 & 0.4553 \\
\hline

\multirow{4}{*}{\makecell{\color{nicered}NHP Loss\\\color{nicered}Context-Free}}
& ArXiv    & 0.8054 & \textbf{0.8438} & 0.3492 & 0.6995 & 0.9977 & \textbf{1.0000} & 0.5021 \\
& News     & 0.7678 & 0.8134 & 0.6261 & 0.3333 & 0.9909 & 0.1565 & 0.4284 \\
& YouTube  & 0.6705 & 0.3333 & 0.5277 & 0.3333 & 0.9362 & 0.3347 & 0.4327 \\
& Combined & 0.6787 & 0.8158 & 0.6202 & 0.8368 & 0.9476 & 0.2000 & 0.4638 \\
\hline

\multirow{4}{*}{\makecell{\color{nicegreen}NSP Context-\\\color{nicegreen}Aware Data Aug.}}
& ArXiv    & 0.8170 & 0.7600 & 0.4973 & 0.8371 & \textbf{1.0000} & 0.1728 & 0.4687 \\
& News     & 0.8095 & 0.6230 & 0.4638 & 0.8173 & 0.9909 & 0.2558 & 0.0720 \\
& YouTube  & 0.7123 & 0.7733 & 0.6609 & 0.8291 & 0.9932 & 0.2585 & 0.4588 \\
& Combined & 0.7306 & 0.8377 & 0.5737 & 0.8236 & 0.9089 & 0.5061 & 0.4386 \\
\hline

\multirow{4}{*}{\makecell{\color{nicegreen}NSP Context-\\\color{nicegreen}Free Data Aug.}}
& ArXiv    & 0.8012 & 0.5732 & 0.7246 & 0.7963 & 0.9977 & 0.1415 & 0.3090 \\
& News     & 0.8125 & 0.8018 & 0.4850 & 0.8522 & 0.9954 & 0.7932 & \textbf{0.5034} \\
& YouTube  & 0.7829 & 0.8032 & 0.5242 & 0.8293 & \textbf{1.0000} & 0.1959 & 0.4466 \\
& Combined & 0.7192 & 0.8490 & 0.6084 & 0.8610 & 0.9863 & 0.3429 & 0.4741 \\
\hline

\multirow{4}{*}{\makecell{\color{nicegreen}NHP Context-\\\color{nicegreen}Aware Data Aug.}}
& ArXiv    & 0.7412 & \textbf{0.8438} & 0.6048 & 0.8227 & \textbf{1.0000} & 0.0177 & 0.4396 \\
& News     & 0.7833 & 0.8117 & 0.3593 & 0.7065 & 0.9863 & 0.4163 & 0.4730 \\
& YouTube  & 0.7005 & 0.8374 & 0.6299 & 0.8469 & 0.7585 & 0.1660 & 0.4368 \\
& Combined & 0.7227 & 0.8116 & 0.5852 & \textbf{0.8644} & 0.9066 & 0.3293 & 0.4241 \\
\hline

\multirow{4}{*}{\makecell{\color{nicegreen}NHP Context-\\\color{nicegreen}Free Data Aug.}}
& ArXiv    & \textbf{0.8178} & 0.6633 & 0.7397 & 0.7887 & 0.9863 & 0.4068 & 0.4464 \\
& News     & 0.7899 & 0.7752 & 0.6354 & 0.8184 & 0.9932 & 0.2898 & 0.4456 \\
& YouTube  & 0.7631 & 0.5725 & 0.6589 & 0.3333 & 0.9977 & 0.3320 & 0.4497 \\
& Combined & 0.6731 & 0.8269 & \textbf{0.7928} & 0.8492 & 0.9841 & 0.0939 & 0.4151 \\

\hline\hline
\multirow{4}{*}{\makecell{\color{niceblue}NTP Synonym\\\color{niceblue}Baseline Fine-Tuned}}
& ArXiv    & 0.7750 & 0.7315 & 0.6385 & 0.8389 & 0.8975 & 0.2571 & 0.3728 \\
& News     & 0.6154 & 0.3405 & 0.4509 & 0.8285 & 0.8975 & 0.2381 & 0.4350 \\
& YouTube  & 0.7221 & 0.4800 & 0.5847 & 0.7952 & 0.9954 & 0.1211 & 0.4358 \\
& Combined & 0.7843 & 0.7428 & 0.5261 & 0.8530 & 0.9841 & 0.3837 & 0.4803 \\
\hline

\multirow{4}{*}{\makecell{\color{niceblue}NTP Hypernym\\\color{niceblue}Baseline Fine-Tuned}}
& ArXiv    & 0.7872 & 0.6711 & 0.6681 & 0.8406 & 0.9932 & 0.0422 & 0.4307 \\
& News     & 0.8176 & 0.8251 & 0.5352 & 0.8218 & 0.8136 & 0.8136 & 0.4483 \\
& YouTube  & 0.7745 & 0.8354 & 0.7098 & 0.8188 & 1.0000 & 0.9537 & 0.4867 \\
& Combined & 0.6946 & 0.6239 & 0.6849 & \textbf{0.8644} & 0.9954 & 0.7279 & 0.4499 \\
\hline
\makecell{\color{niceblue}Base Model\\\color{niceblue}Fine-Tuned} & - & 0.7167 & 0.3333 & 0.4010 & 0.7718 & 1.0000 & 0.0367 & 0.3875 \\
\hline
\makecell{\color{niceblue}Base Model} & - & 0.2000 & 0.3333 & 0.3333 & 0.3333 & 0.000 & 0.0000 & 0.0769 \\
\hline
\end{tabular}}
\end{table*}

\begin{table*}[h!]
\centering
\small
\caption{\textbf{Downstream fine-tuned AUC-PR scores across seven benchmarks.}
Precision is reported for \textsc{Empathetic Dialogues} (EMO), \textsc{GLUE}, \textsc{Hate Speech} (HATE), \textsc{SNLI}, \textsc{SpamAssassin} (SPAM), \textsc{Fake News} (FAKE), \textsc{Logical Fallacy} (LOG).}
\label{tab:downstream-all-aupr}
\resizebox{\textwidth}{!}{
\begin{tabular}{ll|rrrrrrr}
\hline\hline
\textbf{Variant} & \textbf{Domain} & \textbf{EMO} & \textbf{GLUE} & \textbf{HATE} & \textbf{SNLI} & \textbf{SPAM} & \textbf{FAKE} & \textbf{LOG} \\
\hline\hline

\multirow{4}{*}{\makecell{\color{nicered}NSP Loss\\\color{nicered}Context-Aware}}
& ArXiv    & 0.8536 & 0.8321 & 0.6182 & 0.8829 & 0.9987 & 0.5256 & 0.4452 \\
\cdashline{2-9}
& News     & 0.8503 & 0.8534 & 0.7015 & 0.3444 & 0.9994 & 0.4767 & 0.5466 \\
\cdashline{2-9}
& YouTube  & 0.8603 & 0.9084 & 0.6871 & 0.6679 & 0.9794 & 0.5082 & 0.5540 \\
\cdashline{2-9}
& Combined & 0.8602 & 0.3417 & 0.7375 & 0.4181 & 0.9957 & 0.6259 & 0.5401 \\

\hline
\multirow{4}{*}{\makecell{\color{nicered}NSP Loss\\\color{nicered}Context-Free}}
& ArXiv    & 0.8347 & 0.9101 & 0.6479 & 0.5588 & 0.9926 & 0.5207 & 0.4835 \\
\cdashline{2-9}
& News     & 0.8021 & 0.8967 & 0.6771 & 0.3444 & 0.9937 & 0.4711 & 0.5350 \\
\cdashline{2-9}
& YouTube  & 0.8579 & 0.8490 & 0.6512 & 0.3577 & 0.9953 & 0.5666 & 0.5540 \\
\cdashline{2-9}
& Combined & 0.7443 & 0.7159 & 0.3818 & 0.3536 & 0.9960 & 0.4075 & 0.1166 \\

\hline
\multirow{4}{*}{\makecell{\color{nicered}NHP Loss\\\color{nicered}Context-Aware}}
& ArXiv    & 0.8414 & 0.9120 & 0.5095 & 0.7862 & \textbf{0.9999} & 0.6118 & 0.5654 \\
\cdashline{2-9}
& News     & 0.8340 & 0.9043 & 0.5486 & 0.8825 & 0.9980 & 0.5130 & 0.5652 \\
\cdashline{2-9}
& YouTube  & 0.8155 & 0.3578 & 0.6258 & 0.8798 & 0.9955 & 0.4323 & 0.5191 \\
\cdashline{2-9}
& Combined & 0.8276 & 0.8832 & 0.7041 & 0.8145 & 0.9986 & 0.4921 & 0.5520 \\

\hline
\multirow{4}{*}{\makecell{\color{nicered}NHP Loss\\\color{nicered}Context-Free}}
& ArXiv    & 0.8634 & 0.9120 & 0.5095 & 0.7862 & \textbf{0.9999} & 0.4550 & 0.5654 \\
\cdashline{2-9}
& News     & 0.8284 & 0.8728 & 0.6735 & 0.3749 & 0.9974 & 0.5161 & 0.5373 \\
\cdashline{2-9}
& YouTube  & 0.8037 & 0.6069 & 0.6427 & 0.3977 & 0.9955 & 0.4529 & 0.5233 \\
\cdashline{2-9}
& Combined & 0.7939 & 0.8885 & 0.6955 & 0.8884 & 0.9879 & 0.5667 & 0.5831 \\

\hline
\multirow{4}{*}{\makecell{\color{nicegreen}NSP Context-\\\color{nicegreen}Aware Data Aug.}}
& ArXiv    & 0.8619 & 0.8564 & 0.6458 & 0.9111 & \textbf{0.9999} & 0.5265 & 0.5265 \\
\cdashline{2-9}
& News     & 0.8503 & 0.7816 & 0.6518 & 0.8889 & 0.9994 & 0.4924 & 0.1007 \\
\cdashline{2-9}
& YouTube  & 0.7730 & 0.8628 & 0.6920 & 0.8976 & 0.9958 & 0.5176 & 0.5473 \\
\cdashline{2-9}
& Combined & 0.8079 & 0.9031 & 0.6560 & 0.8885 & 0.9916 & 0.5660 & 0.5542 \\

\hline
\multirow{4}{*}{\makecell{\color{nicegreen}NSP Context-\\\color{nicegreen}Free Data Aug.}}
& ArXiv    & 0.8608 & 0.6327 & 0.7132 & 0.8721 & 0.9993 & 0.5157 & 0.3754 \\
\cdashline{2-9}
& News     & 0.8604 & 0.8746 & 0.5379 & 0.9153 & 0.9995 & 0.4974 & \textbf{0.6211} \\
\cdashline{2-9}
& YouTube  & 0.8388 & 0.8656 & 0.6197 & 0.9008 & \textbf{0.9999} & 0.4909 & 0.5700 \\
\cdashline{2-9}
& Combined & 0.8148 & \textbf{0.9175} & 0.6734 & 0.9171 & 0.9986 & 0.4538 & 0.5675 \\

\hline
\multirow{4}{*}{\makecell{\color{nicegreen}NHP Context-\\\color{nicegreen}Aware Data Aug.}}
& ArXiv    & 0.8048 & 0.3009 & 0.6506 & 0.8861 & 0.9987 & 0.6118 & 0.5334 \\
\cdashline{2-9}
& News     & 0.8340 & 0.8846 & 0.6180 & 0.7986 & 0.9997 & 0.5130 & 0.5501 \\
\cdashline{2-9}
& YouTube  & 0.7953 & 0.9109 & 0.7297 & 0.9127 & 0.9956 & 0.5321 & 0.5129 \\
\cdashline{2-9}
& Combined & 0.8273 & 0.8987 & 0.6768 & 0.9192 & 0.9972 & 0.5591 & 0.4992 \\

\hline
\multirow{4}{*}{\makecell{\color{nicegreen}NHP Context-\\\color{nicegreen}Free Data Aug.}}
& ArXiv    & 0.8788 & 0.7568 & 0.7500 & 0.8646 & 0.9995 & 0.5046 & 0.6078 \\
\cdashline{2-9}
& News     & 0.8375 & 0.8477 & 0.6778 & 0.8874 & 0.9961 & 0.4704 & 0.5713 \\
\cdashline{2-9}
& YouTube  & 0.8271 & 0.6069 & 0.6897 & 0.3977 & \textbf{0.9999} & 0.4529 & 0.5233 \\
\cdashline{2-9}
& Combined & 0.8150 & 0.8986 & \textbf{0.7528} & 0.9176 & 0.9969 & 0.4781 & 0.5272 \\

\hline\hline
\multirow{4}{*}{\makecell{\color{niceblue}NTP Synonym\\\color{niceblue}Baseline Fine-Tuned}}
& ArXiv    & 0.8176 & 0.8255 & 0.6714 & 0.9078 & 0.9828 & 0.5384 & 0.5030 \\
\cdashline{2-9}
& News     & 0.7643 & 0.3707 & 0.5535 & 0.9047 & 0.9925 & 0.5994 & 0.5293 \\
\cdashline{2-9}
& YouTube  & 0.8237 & 0.5457 & 0.6541 & 0.8799 & 0.9987 & 0.4934 & 0.5205 \\
\cdashline{2-9}
& Combined & 0.8265 & 0.8517 & 0.6254 & 0.9130 & 0.9882 & 0.5210 & 0.5802 \\

\hline
\multirow{4}{*}{\makecell{\color{niceblue}NTP Hypernym\\\color{niceblue}Baseline Fine-Tuned}}
& ArXiv    & 0.8285 & 0.8127 & 0.7197 & 0.9056 & 0.9985 & \textbf{0.6426} & 0.5143 \\
\cdashline{2-9}
& News     & \textbf{0.8645} & 0.9018 & 0.6489 & 0.8560 & 0.9997 & 0.5219 & 0.5401 \\
\cdashline{2-9}
& YouTube  & 0.8322 & 0.9005 & 0.7445 & 0.9019 & 0.9985 & 0.4934 & 0.5614 \\
\cdashline{2-9}
& Combined & 0.7904 & 0.7005 & 0.7199 & \textbf{0.9324} & 0.9984 & 0.5313 & 0.5459 \\

\hline
\makecell{\color{niceblue}Base Model\\\color{niceblue}Fine-Tuned} & - & 0.7904 & 0.3306 & 0.5817 & 0.8489 & 0.9953 & 0.5439 & 0.5110 \\
\hline
\makecell{\color{niceblue}Base Model} & - & 0.6000 & 0.6667 & 0.6667 & 0.6667 & 0.8424 & 0.7162 & 0.5385 \\
\hline
\end{tabular}}
\end{table*}

\end{document}